\documentclass[11pt]{article}

\usepackage[preprint]{acl}
\hypersetup{pdftitle={Poolwise: Evaluating Reward Models for Selection across Prompts},pdfauthor={Mateusz Modrzejewski and Tomasz Radzikowski}}

\usepackage{times}
\usepackage{latexsym}
\usepackage[T1]{fontenc}
\usepackage[utf8]{inputenc}
\usepackage{microtype}
\usepackage{inconsolata}
\usepackage{graphicx}
\usepackage{booktabs}
\usepackage{amsmath}
\usepackage{array}
\renewcommand{\topfraction}{0.95}
\renewcommand{\bottomfraction}{0.9}
\renewcommand{\textfraction}{0.03}
\renewcommand{\floatpagefraction}{0.85}
\renewcommand{\dbltopfraction}{0.95}
\renewcommand{\dblfloatpagefraction}{0.8}
\newcommand{\ci}[3]{\begin{tabular}[t]{@{}r@{}}#1\\[-2.5pt]{\scriptsize[#2,\,#3]}\end{tabular}}

\title{Poolwise: Evaluating Reward Models for Selection across Prompts}

\author{Mateusz Modrzejewski \\ Warsaw University of Technology \\ \texttt{mateusz.modrzejewski@pw.edu.pl}
  \And Tomasz Radzikowski \\ Warsaw University of Technology \\ \texttt{tomasz.radzikowski@pw.edu.pl}}

\begin{document}
\maketitle

\begin{abstract}
Reward models are used to select training data, but common benchmarks compare responses only within the same
prompt. Selecting data across prompts also depends on how scores compare between prompts. We introduce Poolwise, a
protocol that selects a fixed fraction of a response pool by score, measures the selected set's quality using
external labels, and reports within-prompt ranking separately. In matched experiments on two rating sets and two
model sizes, training each prompt's mean score to match its average response rating improves selection over
permuted or omitted mean targets in every seed at 30 percent retention. In grouped HelpSteer experiments, the model
with the highest within-prompt accuracy selects a lower-rated set than the correct-mean model. Subtracting prompt
means estimated from separate reference responses preserves local rankings but lowers selected helpfulness for six
open reward models and two closed judges on HelpSteer. Poolwise provides a direct way to compare reward models and
score transformations for data selection.
\end{abstract}

\section{Introduction}
\label{sec:intro}

Reward models assign scores to language-model responses and guide preference optimization and data selection
\citep{stiennon2020learning,ouyang2022training,yu2025rip}. Choosing the best response to one prompt requires comparing
only its candidate responses. Filtering a dataset that spans many prompts also requires comparing scores across
prompts. Adding the same constant to all scores for one prompt preserves its response ranking but can change which
responses are selected from the full dataset.

Benchmarks such as RewardBench assess whether a model ranks responses to the same prompt in the order preferred by
humans \citep{lambert2024rewardbench,malik2025rewardbench}. The Bradley--Terry training objective also depends only on
score differences within a prompt, leaving a shared prompt-dependent offset unconstrained
\citep{wang2024helpsteer2,hasanaliyev2026expected,lamparth2026reward}. Previous work studies prompt-baseline
subtraction \citep{wang2025template}, score comparability across prompts using satisfaction labels
\citep{peng2026wildreward}, and differences between overall judge agreement and within-prompt decisions
\citep{landesberg2026judge}. We ask how this prompt-level score information affects the quality of responses selected
across prompts. We also ask whether training can improve that selection.

We introduce Poolwise, a protocol for evaluating this selection decision. Given a scored response pool, external
quality labels and the number of responses to retain, Poolwise measures the quality of the selected set and reports
within-prompt ranking separately. We use it to study two interventions: subtracting each prompt's mean score on
separately generated reference responses, and training otherwise matched reward models with correct, permuted or
omitted mean-rating targets. The first changes scores while preserving each prompt's response ranking. The second
tests whether supervision on observed mean response ratings improves selection.

\begin{figure*}[t]
\centering
\includegraphics[width=\textwidth]{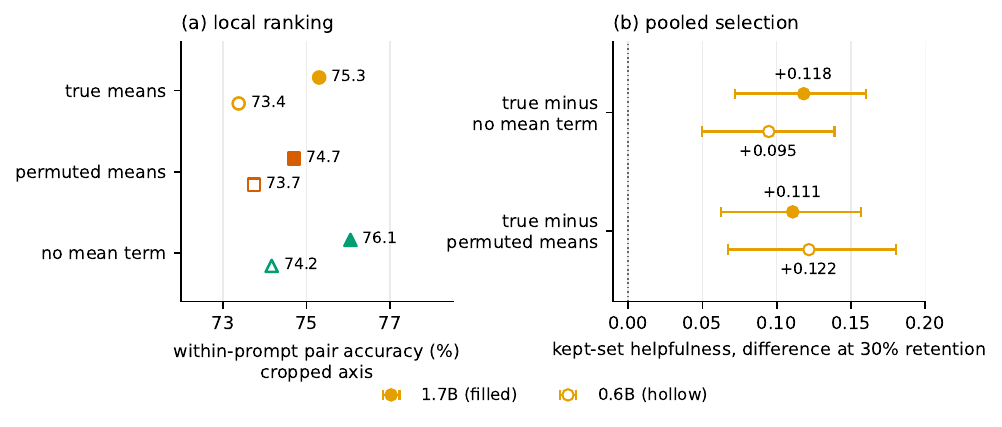}
\caption{Correct mean-rating targets improve pooled selection. Qwen3 models (0.6B and 1.7B, five matched seeds per
condition) are evaluated on 1,123 held-out HelpSteer training-split prompts. (a) Mean within-prompt pair accuracy,
on a cropped axis. (b) Improvement in selected mean helpfulness at 30 percent retention, with Bonferroni-adjusted
95 percent crossed prompt-and-seed bootstrap intervals. At both sizes, the no-mean model has the highest pair
accuracy but lower selected helpfulness than the correct-mean model.}
\label{fig:early}
\end{figure*}

Correct mean-rating targets improve response selection in matched training across two rating sets and two model
sizes at 30 percent retention. On held-out HelpSteer groups, the model with the highest within-prompt accuracy
selects lower-rated responses than the correct-mean model (Figure~\ref{fig:early}). Reference centering provides a
complementary test: it preserves every within-prompt ranking but reduces selected helpfulness for all six core
reward models on HelpSteer, by 0.07 to 0.33 points on the 0 to 4 scale. Two closed judges show the same selection
loss on HelpSteer.

This paper makes three contributions. First, we introduce Poolwise, a protocol for comparing reward models and
score transformations by the quality of the responses they select across prompts. Second, we find that reference
centering can lower selected quality without changing within-prompt rankings, and we explain the loss through the
responses it replaces. Third, we find that correctly assigned mean-rating targets improve pooled selection in
matched training.

\section{Pooled response selection}
\label{sec:task}

A reward model $r(x, y)$ maps a prompt $x$ and a response $y$ to a scalar. Let $r'(x, y) = r(x, y) + c(x)$ for
any function $c$ of the prompt alone. For every pair of responses to one prompt, $r'$ and $r$ differ by the same
constant, so every within-prompt comparison, best-of-$n$ choice and pairwise accuracy is identical for the two.
For a decision that pools responses across prompts, $c$ moves whole prompts up or down the common scale and can
change the outcome. Table~\ref{tab:operations} in the appendix summarizes these cases.

On a candidate pool of $N$ responses with an external outcome $u_i$ per response, a scorer keeps the $K =
\mathrm{round}(qN)$ responses with the highest scores, and the quality of its decision is the mean outcome of the
kept set, $U_q = \frac{1}{K}\sum_{i \in S_q} u_i$. We keep 30 percent of responses in the primary analysis and vary
retention from 10 to 70 percent in a sensitivity analysis. For rating outcomes, $U_q$ is the mean observed rating of
the selected responses. For binary checks, it is their pass rate. The pool mean is the expected quality under
uniform random selection and serves as the baseline. Sorting by the observed labels gives an upper bound on
observed-label quality rather than a ceiling on true quality.

Two further quantities explain a change in $U_q$. For two kept sets $A$ and $B$ of the same size,
\begin{equation}
U(B) - U(A) = \frac{|B \setminus A|}{K}\,\big(\bar u_{B \setminus A} - \bar u_{A \setminus B}\big),
\end{equation}
so a difference decomposes into the fraction of the kept set exchanged and the quality of the promoted and
displaced responses. Fix a scorer and rank each prompt's candidates by its scores. An allocation retains the first
$k_p$ candidates from each prompt, with $\sum_p k_p = K$. The highest observed-label quality achievable under these
constraints is the prefix oracle. Allocation headroom is the gap between the scorer's selected quality and the
prefix oracle, what a better distribution of slots across prompts could recover without changing any local order.
Ordering headroom is the gap between the prefix oracle and the unrestricted label bound. A shared prompt shift leaves
the prefix oracle unchanged, so its effect on selected quality is exactly a change in allocation headroom. Because
boundary ties can make some prefix allocations unreachable through prompt offsets, we report tie counts and treat
the oracle as an upper bound.

\section{Poolwise: the proposed protocol and instruments}
\label{sec:method}

\paragraph{The protocol.}
Poolwise takes a candidate table (prompt, response, source, split role), one or more score tables, an outcome table
naming its quality criterion, the number of responses to keep, and a label-independent tie rule. It returns the
selected response identities and $U_q$ with a paired percentile-bootstrap interval over prompts. Selection is
recomputed inside every resample. It reports a companion local measure: within-prompt pair accuracy on pairs whose
outcomes differ, or the mean outcome of each prompt's top-scored response. It also records prompt coverage, kept
length, ties, missingness and provenance. We will release the implementation and a report that regenerates the
paper's primary table from the declared inputs.

\paragraph{Reference centering.}
The fixed-scorer intervention subtracts from a response's score the mean score of $m$ reference responses to the
same prompt, sampled separately from the candidates, so that no candidate enters its own baseline. We use
references from two instruction models: twelve per prompt on HelpSteer and RoSE and six on IFEval. We also report
leave-one-out centering on the prompt's other rated responses, a partial weight chosen on development prompts, and
an empirical-Bayes shrinkage of the reference mean. The centered score keeps every within-prompt order. The
reference mean depends on the generator and decoding settings that produced the references, which we call the
reference policy. Section~\ref{sec:sensitivity} varies it.

\paragraph{Matched training targets.}
For a group of $n$ responses to one prompt with rewards $r_j$, ratings $h_j$ and errors $e_j = r_j - h_j$, write
$L_W = \frac{1}{n}\sum_j (e_j - \bar e)^2$ and $L_B(t) = (\bar r - t)^2$. The pointwise regression loss on the group
decomposes as $L_W + L_B(\bar h)$: a within-prompt error term and a squared error between the mean reward and the
mean rating. Permuting the means across prompts gives $L_W + L_B(\bar h_{\pi(p)})$, and dropping the mean term gives
$L_W$ alone. The three arms share the base model, the training groups, the optimizer, the number of optimizer
updates and the seeds, and differ in whether the loss uses the correctly assigned mean target, a permuted target,
or no mean term. For pairs, $L_W$ is the squared difference of the two errors over four, so an arm trained on the
within-pair rating gap alone is the omitted-mean arm. Bradley--Terry training and its rating-margin variant provide
pairwise baselines.

\paragraph{Diagnostics.}
Three statistics describe where a score's variation lies (Section~\ref{sec:cluster}). The between-prompt variance
share measures the fraction of score variation that lies between prompts. We estimate it as an intraclass
correlation over the rated responses and compare it with the same statistic for the ratings. The sibling-rating
regression fits a response's standardized score on its own rating and on the rating of a sibling response to the
same prompt. A positive sibling coefficient means that the sibling rating is associated with the score after
conditioning on the response's own rating. We rescore each response under its own prompt, one fixed generic
instruction, wrong prompts, and three paraphrases. For each condition, we report the between-prompt variance share.
For wrong prompts, we also report the score decrease in standard deviations of the original scores. Unless stated
otherwise, intervals are 95 percent percentile-bootstrap intervals over prompts. The grouped HelpSteer training
experiment uses the crossed prompt-and-seed bootstrap described in Appendix~\ref{app:format}.

\section{Experimental setup}
\label{sec:setup}

\paragraph{Scorers.}
The core analysis uses six 7B to 8B reward models: three trained on pairwise preferences with one output logit
(Skywork-Reward-V2-Qwen3-8B, Skywork-Reward-Llama-3.1-8B-v0.2, Llama-3.1-Tulu-3-8B-RM) and three with attribute
heads trained on pointwise ratings (ArmoRM-Llama3-8B, URM-LLaMa-3.1-8B and the Mistral-7B attribute model of DPA),
scored on their helpfulness head. We also evaluate two model families at several sizes, three larger reward models,
three open generative judges, two closed judges, and WildReward-8B \citep{peng2026wildreward}. The open judges are
scored by the expected helpfulness digit. The closed judges, Claude Sonnet 5.5 and Claude Haiku 5.5, are scored by
the mean of ten sampled rating digits. WildReward-8B is an ordinal-regression model trained on interaction-derived
satisfaction labels and scored by its documented procedure. Table~\ref{tab:models} in the appendix lists revisions,
identifiers, training signals and documented data exposure. The matched arms fine-tune Qwen3 at 0.6B and 1.7B in
two settings: on the 6,667 untied HelpSteer2 training pairs with equal numbers of hyperparameter-search trials for
the two main objectives and the regression configuration for the control arms (Appendix~\ref{app:format}), and on
complete four-response groups of HelpSteer's training split with one configuration for all three conditions
(Section~\ref{sec:supervision}). Every arm has five seeds.

\paragraph{Panels and criteria.}
Table~\ref{tab:panels} lists the three selection panels. The primary panel is HelpSteer's validation split
\citep{wang2023helpsteer}, 378 single-turn prompts with four responses each, rated independently by one annotator
per response on a 0 to 4 helpfulness scale. Its between-prompt rating share is 0.26, so prompt-level quality is
part of the criterion. Two external panels change domain, source and criterion together, so they are transfer
tests rather than criterion experiments. RoSE \citep{liu2023rose} contributes 500 CNNDM test articles with the
eight systems common to its validation and test panels, scored on normalized atomic-content-unit recall, with
every choice made on its validation panel and the test panel evaluated once. IFEval \citep{zhou2023ifeval}
contributes 541 prompts with six generated candidates each, scored by the official strict and loose compliance
checkers at a pinned commit, with all-pass and all-fail prompts kept. HelpSteer2 validation \citep{wang2024helpsteer2}
serves the clustering diagnostics and the threshold-transport check. Its between-prompt rating share is near zero.

\begin{table}[t]
\centering
\scriptsize
\setlength{\tabcolsep}{2pt}
\begin{tabular}{lrrrl}
\toprule
Panel & Prompts & Pool & $K$ & Criterion \\
\midrule
HelpSteer validation & 378 & 1,512 & 454 & helpfulness, 0 to 4 \\
RoSE CNNDM test & 500 & 4,000 & 1,200 & ACU recall, 0 to 1 \\
IFEval & 541 & 3,246 & 974 & strict compliance \\
\bottomrule
\end{tabular}
\caption{Selection panels at 30 percent retention. Every candidate is scored by all six core models. References
are sampled separately from the candidates with Qwen3-1.7B and Llama-3.2-3B-Instruct: six from each per prompt on
HelpSteer and RoSE, three from each on IFEval.}
\label{tab:panels}
\end{table}

\paragraph{References and splits.}
References are sampled at temperature 0.7 and top-p 0.9 under each generator's chat template, kept without
filtering, and never selected by any label. They are two to three times longer than the HelpSteer responses and
several times longer than the RoSE summaries (appendix length columns). Development and evaluation roles use two
prompt folds with a fixed seed. We choose models and partial-centering weights and fit calibrators on one fold,
evaluate them on the other, and combine the held-out results weighted by the number of responses kept. The retention
count is $\mathrm{round}(0.30 N)$ throughout.

\paragraph{Scoring validation.}
Across twelve RewardBench or RewardBench 2 comparisons covering seven models, our reproduced scores differ from the
published scores by at most 1.0 point (median absolute gap 0.35). ArmoRM's published benchmarks use its combined
output. We reproduce that output within 1.2 points on RewardBench and 4.8 points below the published score on
RewardBench 2. Our main analyses use its helpfulness head. Rescoring three models on a second GPU type in fp32
instead of bf16 gives rewards that correlate with the originals at 0.9996 or higher.

\section{Mean-target supervision improves pooled selection}
\label{sec:supervision}

We trained the three conditions of Section~\ref{sec:method}, correct, permuted and omitted prompt-mean targets, on
complete four-response groups from 5,239 prompts in HelpSteer's training split. At each size, we used one
configuration and five seeds with matched initialization and data order, then evaluated once on 1,123 held-out test
prompts (Figure~\ref{fig:early}, Table~\ref{tab:ea2}). At 1.7B and 30 percent retention, the correct-mean arm selects
a set rated 0.118 [0.072, 0.160] points above the omitted-mean arm (no mean term) and 0.111 [0.062, 0.157] above the
permuted-mean arm, with Bonferroni-adjusted intervals from a crossed prompt-and-seed bootstrap and a positive
difference in every seed pair. At 0.6B the same contrasts are 0.095 [0.050, 0.139] and 0.122 [0.067, 0.180], again
positive in every seed pair. The omitted-mean arm has the highest within-prompt pair accuracy at both sizes (0.761
against 0.753 at 1.7B, 0.742 against 0.734 at 0.6B), yet selects lower-rated responses than the correct-mean arm.
The correct-mean arm also outperforms both control arms on HelpSteer validation, under raw scoring at both sizes
and, at 1.7B, under centering with either reference pool of Section~\ref{sec:sensitivity}, with a positive difference
in every seed pair.

The correct-mean arm also outperforms both control arms when trained on HelpSteer2 preference pairs, whose ratings
carry almost no prompt-level component, with the pairwise objectives as further baselines (Table~\ref{tab:primary},
Figure~\ref{fig:supervision}). At 1.7B its selected set is rated 0.078 [0.039, 0.136] points higher than the
permuted-mean arm's and 0.060 [0.028, 0.103] higher than the omitted-mean arm's, averaged over five matched seeds
with a positive difference in every seed pair, and 0.180 [0.124, 0.244] higher than the Bradley--Terry arm's. At
1.7B, within-prompt pair accuracy is 0.681 for correct means, 0.664 for permuted means and 0.650 with no mean term,
so these separately trained networks also differ in local ranking, by up to 3.1 percentage points. The advantage
over the permuted-mean arm is larger in pooled selection than in best-of-four selection: 0.078 against 0.021
[$-0.012$, 0.053] rating points (Table~\ref{tab:localcontrols}). The two arms order 0.160 of the within-prompt pairs
differently, within the 0.15 to 0.19 observed between retrains of one configuration, and the permuted arm's selected
set exchanges 0.27 to 0.34 of its responses with the correct-mean arm's per seed, with the promoted responses rated
2.83 to 2.98 against 3.06 to 3.27 for the displaced ones.

\section{Reference centering preserves local rankings but lowers selected quality}
\label{sec:centering}

\begin{figure*}[t]
\centering
\includegraphics[width=\textwidth]{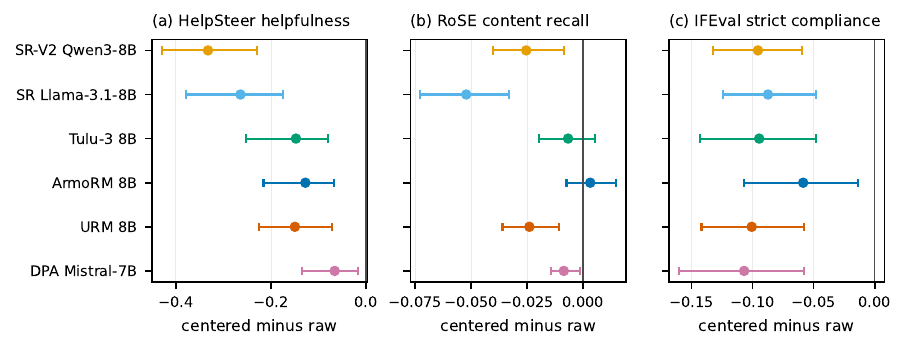}
\caption{Change in selected quality after reference centering at 30 percent retention, per core model: (a) HelpSteer
validation mean helpfulness (0 to 4), (b) RoSE test atomic-content-unit recall (0 to 1), (c) IFEval strict
compliance (0 to 1). Values are centered minus raw, so negative values favor raw scores. Bars are paired 95 percent
prompt-bootstrap intervals.}
\label{fig:intervention}
\end{figure*}

\begin{figure}[t]
\centering
\includegraphics[width=\columnwidth]{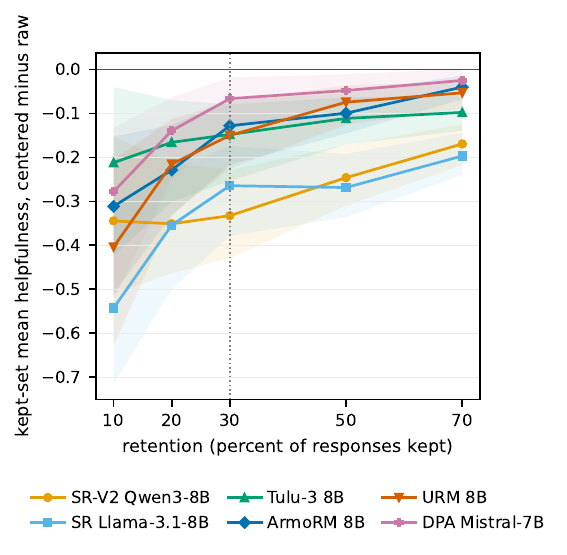}
\caption{Centering effect across retention on HelpSteer validation for the six core models, with twelve references
per prompt. Values are changes in selected mean helpfulness, centered minus raw. Shading shows paired 95 percent
prompt-bootstrap intervals. The dotted line marks 30 percent retention.}
\label{fig:retention}
\end{figure}

On HelpSteer, centered-minus-raw changes in the selected set's mean helpfulness range from $-0.066$ [$-0.134$,
$-0.018$] for DPA to $-0.333$ [$-0.430$, $-0.229$] for Skywork-Reward-V2-Qwen3-8B, with every interval below zero
(Figure~\ref{fig:intervention}a). The loss holds at every retention from 10 to 70 percent and is largest at 10
percent for five of the six (Figure~\ref{fig:retention}). Raw scores select responses with mean helpfulness of 3.20
to 3.33, compared with 2.83 under random selection and an observed-label upper bound of 3.73.

Centering exchanges 23 to 50 percent of the 454 selected responses, and the promoted responses are rated 2.65 to
2.92 against 3.15 to 3.31 for the displaced ones (Table~\ref{tab:exchange}). Most of each scorer's shortfall from the
observed-label bound is allocation headroom (0.36 to 0.48 rating points), compared with 0.035 to 0.055 points of
ordering headroom. Because centering preserves each prompt's ranking, its loss in selected helpfulness adds entirely
to allocation headroom. In this panel centering reallocates selections away from prompts with higher observed
candidate quality: of the 454 selected responses, 29 to 102 fewer come from the third of prompts with the best-rated
candidates and 28 to 91 more from the third with the worst.
Leave-one-out centering on the rated siblings and the shrinkage baseline also lower selected helpfulness, and the
partial weight chosen on development prompts is zero for every model but one fold of Tulu-3 (Table~\ref{tab:pool}
reports a separate run of the same sweep). The matched arms lose under centering as well. Scored on the same pool,
the correct-mean arm and retrained control arms select sets rated 0.07 to 0.16 points lower. The correct-mean arm
remains highest, and centering reverses the ordering of the two retrained controls, with the permuted-mean arm
above the omitted-mean arm (Table~\ref{tab:armscentered}).

\begin{table*}[t]
\centering
\small
\resizebox{\textwidth}{!}{
\begin{tabular}{lrrrrrrr}
\toprule
& & & & & \multicolumn{2}{c}{Allocation headroom} & \\
Model & Exchanged & Promoted & Displaced & Helpfulness change & raw & centered & Ordering headroom \\
\midrule
SR-V2-Qwen3-8B & 0.50 & 2.65 & 3.31 & -0.333 & 0.425 & 0.758 & 0.035 \\
SR-Llama-3.1-8B & 0.50 & 2.71 & 3.24 & -0.264 & 0.449 & 0.714 & 0.044 \\
Tulu-3-8B-RM & 0.38 & 2.76 & 3.15 & -0.148 & 0.476 & 0.623 & 0.055 \\
ArmoRM-8B & 0.28 & 2.84 & 3.30 & -0.128 & 0.363 & 0.491 & 0.040 \\
URM-8B & 0.31 & 2.79 & 3.27 & -0.150 & 0.416 & 0.566 & 0.037 \\
DPA-Mistral-7B & 0.23 & 2.92 & 3.21 & -0.066 & 0.386 & 0.452 & 0.035 \\
\bottomrule
\end{tabular}}
\caption{Reference centering on HelpSteer at 30 percent retention. Exchanged is the fraction of selected responses
replaced. Promoted and displaced report the mean helpfulness of responses entering and leaving the selection.
Allocation headroom is the potential gain from reallocating selections while preserving within-prompt order.
Ordering headroom is the remaining gap to the label-sort bound and is unchanged by centering.}
\label{tab:exchange}
\end{table*}

The direction holds on the two external criteria, with one exception. On the RoSE test panel, evaluated once with
every choice fixed on the validation panel, centering lowers selected content recall for five of six models by
point estimate and for four with intervals below zero, by up to 0.052 [0.033, 0.073] for
Skywork-Reward-Llama-3.1-8B-v0.2. ArmoRM's change is $+0.003$ [$-0.007$, $+0.015$] (Figure~\ref{fig:intervention}b).
References drawn from the released systems produce a larger loss than the generated pool for every model. On IFEval,
centering lowers selected strict compliance by 0.059 to 0.107 with every interval below zero, and loose compliance
also decreases for all six models (Figure~\ref{fig:intervention}c). WildReward-8B, trained for cross-query
comparability, selects a set rated 3.24 on HelpSteer and loses 0.165 points under centering, inside the core range.
Reference centering also lowers selected helpfulness for both closed judges on HelpSteer. Claude Sonnet 5.5 and
Claude Haiku 5.5 select sets with mean ratings of 3.23 and 3.26 under raw scoring and lose 0.098 [0.069, 0.222] and
0.215 [0.131, 0.308] points after centering (Table~\ref{tab:closed}).

\section{Reference policy, calibration and model choice}
\label{sec:sensitivity}

The reference mean depends on the reference policy. Relative to the twelve-reference pool, centering on six Qwen
references lowers selected helpfulness by 0.026 to 0.092 points, and centering on six Llama references changes it
by $-0.035$ to $+0.044$, with intervals excluding zero for two of the Qwen comparisons (Table~\ref{tab:composition}).
Six references from a third family, Phi-3.5-mini, produce a smaller loss of selected helpfulness than six from
either other generator, and adding them to the twelve changes the loss by at most 0.035 with every interval
including zero (Table~\ref{tab:thirdfamily}).

The tested reference-based calibrators show no clear held-out improvement over raw scores. Fitting $\hat u = a +
b_W (r - \mu_{\mathrm{ref}}) + b_B \mu_{\mathrm{ref}}$, with $\mu_{\mathrm{ref}}$ the mean reward of the prompt's
generated references, on one fold and keeping the top 30 percent on the other changes the selected set's
helpfulness by $-0.011$ to $+0.018$ relative to raw. The logistic form on the probability of a rating of at least 2,
a centered-only predictor, a prompt-only predictor and a residual prompt head fitted on the development fold also
show no clear held-out improvement over raw scores. The best of them gains 0.024 [$-0.022$, 0.051]
(Table~\ref{tab:calibration}).

Choosing a scorer by held-out selected quality rather than by within-prompt accuracy gains 0.040 [0.007, 0.079]
points on the held-out fold over a nine-scorer panel, a difference that vanishes when the 57 prompts whose four
responses share one rating are dropped (Table~\ref{tab:modelchoice}). Responses tied within one prompt can still
differ in quality from responses to other prompts and affect pooled selection. On the same panel the model with the
highest RewardBench accuracy, the one with the largest wrong-prompt penalty and the one with the lowest
between-prompt share each leave selected helpfulness 0.04 to 0.09 points below the highest value. A raw-score
threshold that keeps 30 percent of the responses on one HelpSteer fold keeps 26 to 35 percent on the other and 46
to 60 percent on HelpSteer2 for five of six models. Under a separately fitted low-helpfulness risk threshold, the
fraction of selected responses rated below 2 stays within five percentage points of its source-fold value for four
of six models (Table~\ref{tab:transport}).

\section{Between-prompt clustering as a diagnostic}
\label{sec:cluster}

On HelpSteer2 validation the six core models have estimated between-prompt variance shares of 0.34 to 0.64,
compared with $-0.01$ for the human ratings on the same prompts, and on HelpSteer all eighteen released scorers have
shares of 0.45 to 0.65 against the observed human share of 0.26 (Tables~\ref{tab:conditions-full} and
\ref{tab:hs1}). Claude Sonnet 5.5 and Claude Haiku 5.5 have shares of 0.46 and 0.50 on HelpSteer2 and 0.60 and 0.55
on HelpSteer, and one fixed instruction in place of the prompt moves their shares by less than 0.06
(Table~\ref{tab:closed}). Substantial clustering by the original prompt remains for every scorer when one fixed
instruction replaces the prompt, and under that instruction the correlation between mean model scores and mean
human ratings across prompts on HelpSteer falls from a range of 0.40 to 0.58 to a range of $-0.03$ to 0.15
(Appendix~\ref{app:support}). Where the ratings carry a prompt level, every model's sibling coefficient is positive,
so the score is associated with the sibling's rating beyond the response's own (Table~\ref{tab:swaphs1}).

These diagnostics do not determine which scorer selects higher-quality responses. A strictly increasing
transformation of the scores preserves every selected set but can change the between-prompt variance share. Adding a
constant to all scores for one prompt preserves within-prompt variance and ranking but can change the selected set.
The selection results show that the prompt-level information in raw scores is useful for this target, so we report
the share, the wrong-prompt penalty and the agreement with prompt-level ratings as explanatory measurements beside
the decision result.

\section{Related work}
\label{sec:related}

\citet{wang2025template} learn a prompt-template offset inside RLHF training and subtract it before policy
optimization. Our fixed-scorer intervention measures what a post-hoc subtraction does to an externally labeled
selection, a different objective, and our matched arms intervene on the mean term inside the loss.
\citet{wang2024hs2preference} compare regression and Bradley--Terry training on the same data and discuss
absolute-score filtering. We vary what the mean term carries within one loss. \citet{peng2026wildreward} evaluate
cross-query comparability on satisfaction-derived outcomes with an ordinal-regression scorer. We evaluate the quality
of a fixed-size selection across prompts and include their released model as a comparator.
\citet{landesberg2026judge} decomposes judge agreement into between-prompt and within-prompt parts and studies
best-of-$n$ decisions. We examine whether within-prompt evaluation predicts the quality of responses selected across
prompts. \citet{hasanaliyev2026expected} predict per-prompt mean rewards from prompt embeddings for routing. We test
how retaining or subtracting the reference mean affects the quality of the selected responses. Calibration of judge
scores to human targets \citep{hashemi2024llmrubric,sahoo2025quantitative,polo2025bridge} and counterfactual prompt
probes \citep{yamamoto2026prism} address related calibration questions. Prompt-mismatch tests through reward gaps
\citep{mao2025information,liu2024rrm,liu2026harve,lu2026robust} detect response-specific mismatch and cancel shared
shifts. We compare between-prompt and within-prompt variation in reward scores with the corresponding variation in
human ratings. The sibling-rating regression and the prompt-only condition are adapted from probes for
audio-language judges of generated music \citep{modrzejewski2026judges,yang2026style}.

\section{Conclusion}

We introduced Poolwise to evaluate reward models by the quality of the responses they select across prompts. Our
controlled comparisons show why this decision needs its own evaluation. Reference centering can lower selected
quality while preserving every within-prompt ranking. Correct mean-rating supervision improves selection at 30
percent retention in matched training on two rating sets and at two model sizes. In the grouped HelpSteer
experiments, the model with the highest within-prompt accuracy selects lower-rated responses than the correct-mean
model.

These findings support evaluating reward models and score transformations on the selections they produce,
alongside within-prompt ranking. Poolwise could also support data curation for fine-tuning small local models, for
example by comparing scorers used to select synthetic responses to product-support questions. We hope it makes
evaluation across prompts a routine part of developing reward models for data selection.

\section*{Limitations}

The released-model centering results use 378 HelpSteer validation prompts and two external panels that change
domain, response source and criterion together. Agreement across the external panels does not isolate the
contribution of domain, source or criterion. The criteria are a mean of ordinal helpfulness ratings, a
content-recall score and a binary compliance check, reported in their own units. Reference centering was tested
with three generator families and up to eighteen references. The reference policy changes the cost. No pool
identifies a pure prompt component. The calibration result concerns the tested family. The closed judges are scored
as sample means of ten digits, which ties 27 and 36 percent of within-prompt pairs.

The pair-trained arms learned from HelpSteer2 preference pairs, whose pair means carry the pair's average quality
while the published between-prompt share of these ratings is near zero, and every control arm reused the
correct-mean arm's configuration, as did the conditions of the grouped experiment. The weights of the 1.7B
permuted-mean and omitted-mean arms were lost after scoring, so their analyses reproduce from the saved score rows
but their scores cannot be regenerated. Retrains with the logged configuration repeat the raw ordering of the three
arms but agree with the originals on only 0.81 to 0.85 of within-prompt pair orders. The grouped experiment repeats
the ordering at two sizes and has archived checkpoints. A preliminary preference-fine-tuning comparison of
raw-filtered and centered-filtered data (Appendix~\ref{app:support}) detected no downstream difference, so benefits
of better selection for a fine-tuned model are not established here.

The human references differ between HelpSteer and HelpSteer2 in response source, rater population and rating
protocol, which this paper cannot separate. The generic-prompt condition replaces the task with a fixed instruction
and may shift the input distribution. Two models have documented exposure to HelpSteer2's validation first turns or
to HelpSteer's ratings, without row-level confirmation.

\section*{Ethical considerations}

The study uses released datasets and open model weights under their terms (Appendix~\ref{app:tables}). We will
release the Poolwise implementation, the code, the scores with source identifiers and the generated prompt sets
under the sources' redistribution terms. Pooled selection changes which prompt groups the selected data represents.

\bibliography{references}

\clearpage\onecolumn\appendix
\renewcommand{\topfraction}{0.95}\renewcommand{\bottomfraction}{0.95}\renewcommand{\textfraction}{0.03}\renewcommand{\floatpagefraction}{0.8}\renewcommand{\dbltopfraction}{0.95}\renewcommand{\dblfloatpagefraction}{0.8}\setcounter{topnumber}{5}\setcounter{bottomnumber}{5}\setcounter{totalnumber}{10}\setcounter{dbltopnumber}{5}

\section{Prompt-shift invariance of pairwise objectives}
\label{app:prop}

Let $r$ be any reward and $c$ any function that is constant across the responses to a prompt, whether it is
computed from the prompt alone or from what the responses to that prompt share, and let $r'(x, y) = r(x, y) +
c(x)$. For every pair $(y^+, y^-)$ under one prompt $x$, $r'(x, y^+) - r'(x, y^-) = r(x, y^+) - r(x, y^-)$, so the
Bradley--Terry loss, any pairwise accuracy, and any within-prompt ranking are identical for $r$ and $r'$. The
between-prompt share of $r'$ is not: it is $(\sigma_a^2 + \sigma_c^2 + 2\,\mathrm{cov}(a, c)) / (\sigma_a^2 +
\sigma_c^2 + 2\,\mathrm{cov}(a, c) + \sigma_e^2)$ for a prompt effect $a$ of $r$, and can take any value in $[0, 1)$
by the choice of $c$. A regression loss on ratings $h$ is not invariant: it penalizes $(r'(x, y) - h(x, y))^2$,
whose expectation over the responses to $x$ contains $(c(x) + a_x - a^h_x)^2$ with $a^h_x$ the ratings' own prompt
effect, so the fitted per-prompt term is pulled toward the ratings'. An additive shift $c(x)$ changes each transplant contrast by $c(x) - c(x')$. The pairwise loss leaves this
contribution unidentified, whereas the regression loss penalizes shifts away from the rated prompt means. Neither
invariance argument predicts the learned mismatch margin on substituted prompt--response pairs.

\section{Experimental details}
\label{app:format}

\paragraph{Crossed bootstrap for the grouped training experiment.}
The contrasts of Section~\ref{sec:supervision} between arms trained with five matched seeds use 2,000 replicates.
Each replicate draws the test prompts with replacement and, independently, the five seeds with replacement, with the
same draw for every condition. Within a replicate, selection is recomputed on the resampled pool for each arm and
seed, the paired difference of kept means is taken per drawn seed, and the differences are averaged over the drawn
seeds. The two declared contrasts per size, true minus omitted and true minus permuted, form the Bonferroni family,
so each interval takes the 1.25 and 98.75 percentiles of the replicates. The point estimate is the mean over the
five seeds of the paired difference.

\paragraph{Grouped training configuration and closed-judge scoring.}
The grouped HelpSteer arms use the pair-trained regression arm's configuration of their size with no further search.
At 1.7B this is a learning rate of $4.1 \times 10^{-6}$, batches of 16 responses (four groups), a warm-up fraction
of 0.0013, a cosine schedule, weight decay 0.1 and two epochs (2,620 updates). At 0.6B it is a learning rate of $1.1
\times 10^{-6}$, batches of 8 responses (two groups), a warm-up fraction of 0.098, a constant schedule, no weight
decay and two epochs (5,240 updates). The closed judges were queried on 2026-11-01 through the provider's batch
interface with the paper's judge template, output limited to eight tokens, a strict one-digit parser, and ten
independent draws per response at the provider's default temperature. The returned model identifiers equal the
requested ones. Of 55,536 requests per model, Claude Haiku 5.5 returned 54,783 parsable digits, 752 outputs that
reached the token limit and one unparsable output, and Claude Sonnet 5.5 returned 55,233 parsable digits, 233 outputs at the limit and 70
refusals, all on HelpSteer2. These are kept as statuses. A response's score is the mean of its parsable draws, and a
prompt is dropped from a selection analysis when one of its candidates has no parsable draw (two HelpSteer validation
prompts).

\begin{table*}[htbp]
\centering
\footnotesize
\resizebox{\textwidth}{!}{
\begin{tabular}{lllllll}
\toprule
Model & Params & Signal & HelpSteer2 & HelpSteer (2023) & Revision / model ID & Licence \\
\midrule
Skywork-Reward-V2-Qwen3-8B & 7.6B & pairwise & not stated & pool & \texttt{6f19fde} & apache-2.0 \\
Skywork-Reward-Llama-3.1-8B-v0.2 & 7.5B & pairwise & yes & no & \texttt{d4117fb} & not stated \\
Llama-3.1-Tulu-3-8B-RM & 7.5B & pairwise & no & no & \texttt{84e607a} & llama3.1 \\
ArmoRM-Llama3-8B-v0.1 & 7.5B & pointwise & not stated & yes & \texttt{eb2676d} & llama3 \\
URM-LLaMa-3.1-8B & 7.5B & pointwise & yes & no & \texttt{5344f0f} & not stated \\
RewardModel-Mistral-7B-for-DPA-v1 & 7.1B & pointwise & not stated & yes & \texttt{86583a7} & not stated \\
Skywork-Reward-V2-Qwen3-0.6B & 0.6B & pairwise & not stated & pool & \texttt{8c14a4e} & apache-2.0 \\
\bottomrule
\end{tabular}}
\caption{Released scorers with parameters, training signal (``both'' is Bradley--Terry combined with regression on the HelpSteer2 attributes, as the card states), documented use of HelpSteer2 and HelpSteer in training (``pool'' means the Skywork-Reward-V2 curation pool, with retention not stated), weights revision or hosted model identifier, and licence as the card states it. The licence of the two NVIDIA models is the NVIDIA Open Model License together with the Llama 3.1 Community License.}
\label{tab:models}
\end{table*}

\begin{table*}[htbp]
\centering
\footnotesize
\setlength{\tabcolsep}{4pt}
\resizebox{\textwidth}{!}{
\begin{tabular}{lll}
\toprule
Condition & Input & Measured contrast \\
\midrule
Original prompt & each response under its own prompt & the between-prompt share \\
Generic prompt & each response under one fixed generic instruction & clustering by the original prompt under a fixed instruction \\
Prompt only & each prompt with one generic reply & association of a fixed-reply score with the prompt means \\
Random transplant & each response under four random other prompts & the mismatch penalty for an unrelated prompt \\
Nearest transplant & the four nearest prompts by TF-IDF or by sentence embedding & the penalty for a lexically or semantically near prompt \\
Length-matched & four random prompts within 10\% of the original prompt's length & the penalty at matched prompt length \\
Related task & three related prompts asking for something else & the penalty for a related task \\
Paraphrase & three rewordings of the original prompt & sensitivity to wording (mean absolute and mean signed shift) \\
\bottomrule
\end{tabular}}
\caption{The prompt-substitution protocol. Each condition is compared with the original-prompt scores.}
\label{tab:conditions-box}
\end{table*}

\begin{table*}[htbp]
\centering
\footnotesize
\begin{tabular}{llrrr}
\toprule
Set, split & Prompts & Observed share & Reliability & Disattenuated share \\
\midrule
HelpSteer2, train, all prompts & 10160 & \ci{0.02}{-0.01}{0.04} & 0.73 (three-rater mean) & \ci{0.02}{-0.01}{0.06} \\
HelpSteer2, train, single-turn & 7207 & \ci{0.01}{-0.02}{0.04} & 0.76 (three-rater mean) & \ci{0.01}{-0.02}{0.05} \\
HelpSteer2, validation, all prompts & 519 & \ci{-0.01}{-0.11}{0.10} & 0.71 (three-rater mean) & \ci{-0.01}{-0.16}{0.14} \\
HelpSteer2, validation, single-turn & 356 & \ci{-0.07}{-0.19}{0.04} & 0.74 (three-rater mean) & \ci{-0.10}{-0.26}{0.06} \\
HelpSteer, validation, all prompts & 378 & \ci{0.26}{0.20}{0.33} & 0.62 (proxy) & \ci{0.42}{0.32}{0.53} \\
HelpSteer, validation, all prompts & 378 & \ci{0.26}{0.20}{0.33} & 0.83 (proxy) & \ci{0.32}{0.24}{0.39} \\
HelpSteer, train, all prompts & 7485 & \ci{0.24}{0.22}{0.25} & 0.62 (proxy) & \ci{0.38}{0.36}{0.41} \\
\bottomrule
\end{tabular}
\caption{Observed and reliability-adjusted human helpfulness shares. The adjusted share is the observed share divided by the rating reliability, which is the three-rater mean on HelpSteer2 and a borrowed proxy on HelpSteer, where one rater rated each response.}
\label{tab:refs}
\end{table*}

\begin{table*}[htbp]
\centering
\footnotesize
\begin{tabular}{lrrrrr}
\toprule
Judge & Top digit 5 & Expected digit & Logit of $P(5)$ & Rank \\
\midrule
Llama-3.1-70B-Instruct & 0.69 & \ci{0.32}{0.22}{0.41} & \ci{0.38}{0.29}{0.47} & \ci{0.46}{0.38}{0.53} \\
Llama-3.1-8B-Instruct & 0.77 & \ci{0.40}{0.30}{0.49} & \ci{0.47}{0.40}{0.55} & \ci{0.52}{0.45}{0.59} \\
Qwen3-8B & 0.55 & \ci{0.50}{0.41}{0.58} & \ci{0.52}{0.45}{0.59} & \ci{0.58}{0.51}{0.64} \\
\bottomrule
\end{tabular}
\caption{Judge score-scale sensitivity on HelpSteer2 validation. The first numeric column is the fraction of responses whose most probable rating is 5, and the last three columns are between-prompt shares on the expected digit, on the logit of the top-digit probability and on ranks.}
\label{tab:scales}
\end{table*}

\begin{table*}[htbp]
\centering
\footnotesize
\begin{tabular}{lrrr}
\toprule
Reward model & Reliability & Share & Share, disattenuated \\
\midrule
URM-LLaMa-3.1-8B & 0.97 & 0.34 & \ci{0.35}{0.25}{0.44} \\
ArmoRM-Llama3-8B-v0.1 & 0.98 & 0.51 & \ci{0.52}{0.42}{0.61} \\
DPA-Mistral-7B-v1 & 0.99 & 0.57 & \ci{0.58}{0.47}{0.69} \\
SR-Llama-3.1-8B-v0.2 & 0.95 & 0.45 & \ci{0.47}{0.38}{0.56} \\
SR-V2-Qwen3-0.6B & 0.97 & 0.60 & \ci{0.62}{0.56}{0.68} \\
SR-V2-Qwen3-8B & 0.97 & 0.53 & \ci{0.54}{0.47}{0.62} \\
Tulu-3-8B-RM & 0.99 & 0.64 & \ci{0.65}{0.57}{0.71} \\
\bottomrule
\end{tabular}
\caption{Paraphrase-based score stability on HelpSteer2 validation. Reliability is one minus the ratio of within-response variance across three prompt paraphrases to total variance. The adjusted share divides the observed between-prompt share by this reliability, as a sensitivity calculation.}
\label{tab:modelrel}
\end{table*}

HelpSteer's validation release has 503 prompts with one to four rated responses. Every HelpSteer analysis keeps the 378 with exactly four, and the 125 others are left out rather than imputed. Of the 519 HelpSteer2 validation prompts, 511 first turns are distinct as written and 509 after lower-casing and whitespace normalization, and grouped folds and held-out splits use the 509. HelpSteer2's training split has 10,161 prompts, of which the 10,160 with two responses enter the share. Multi-turn prompts are passed as their turns with the rated response as the last assistant turn, through each model's chat template. The DPA model uses its fixed string template. Inputs are truncated at 4,096 tokens on the
right. The generic reply of the prompt-only condition is "Thank you for your question. I will do my best to
help.", and the fixed instructions of the generic-prompt condition are "Please respond.", "Write something." and the
empty prompt. The random transplant draws four other prompts per response with a fixed seed. The nearest
transplants take the four prompts with the highest TF-IDF cosine over words or the highest cosine between
sentence embeddings (BAAI/bge-small-en-v1.5), excluding prompts from the same conversation. The length-matched
transplant draws four random prompts within 10\% of the original prompt's word count. The paraphrases and related-task
prompts were written by Qwen3-8B (sampling at temperature 0.7 with a fixed seed) from two fixed instructions. A
variant that repeated the original or was shorter than five words was rejected and regenerated once, after which
50 of the 519 prompts (9.6\%) had no valid paraphrase and were dropped from that condition, and every prompt had
three related tasks. The response probes are ridge, kernel ridge, boosted trees and a multilayer perceptron on two
sentence embeddings (BAAI/bge-small-en-v1.5 and bge-large-en-v1.5) with text features and a topic cluster,
cross-validated by prompt. The three open generative judges were asked for a helpfulness rating from 1 to 5 and scored as the expected
value over the digit tokens. On the HelpSteer2 validation responses the smallest probability mass on the five rating digits is 1.00 for Qwen3-8B, 0.99 for Llama-3.1-8B-Instruct and 0.94 for Llama-3.1-70B-Instruct. On HelpSteer the share of rows under 0.9 is 0.2\%, 8.2\% and 2.3\% on the real rows and 0.3\%, 23.9\% and 5.9\% on transplanted rows, and under the empty reply, a condition not reported for the judges, the 70B judge has 0.87 of its rows under 0.9. The top digit is 5
for 0.55, 0.77 and 0.69 of the validation responses and the median expected value is 4.62, 4.83 and 4.89.

The pair-trained regression and Bradley--Terry arms were each tuned with 24 trials (tree-structured Parzen estimator, median pruner) over the
learning rate (log-uniform from $10^{-6}$ to $5 \times 10^{-5}$), batch size (8, 16, 32), warm-up ratio (0 to 0.1),
schedule (constant, cosine), weight decay (0, 0.01, 0.1) and epochs (1, 2), on within-prompt accuracy over held-out
training pairs grouped by first user turn. The chosen configurations were, for the regression arm, learning rate
$1.1 \times 10^{-6}$, batch 8, warm-up 0.098, constant schedule, no weight decay, two epochs (held-out accuracy
0.718), and for the Bradley--Terry arm, learning rate $5.3 \times 10^{-6}$, batch 16, warm-up 0.054, cosine
schedule, weight decay 0.01, one epoch (0.744). The omitted-mean arm (gap-only in the earlier naming) fits $(d_r - d_h)^2 / 4$ for the two responses of
a pair, with $d_r$ the difference of their rewards and $d_h$ the difference of their ratings, and the shuffled-mean
arm fits the full pointwise loss after the mean rating of each pair has been permuted across pairs. Both use the
regression arm's configuration. The pairs are HelpSteer2's released training preferences: 8,677 pairs, less 1,911 ties and 99 pairs whose first turn opens a validation conversation, leave 6,667 pairs over 6,089 conversations. The chosen response is given by the sign of the preference strength, and it is rated higher on helpfulness in 5,494 pairs, equal in 1,052 and lower in 121. Each arm adds a bias-free linear head to the base model's hidden state at the last non-padding token. All parameters are trained, with fp32 master weights and bf16 autocast. The regression arms fit squared error to the standardized rating of each of the 13,334 responses, the Bradley--Terry arm fits $-\log \sigma(r_c - r_r)$ for the chosen and rejected rewards, the rating-margin arm subtracts the helpfulness gap in raw points inside the sigmoid (zero for 1,173 pairs), and the binarized arm regresses on 1 for the chosen and 0 for the rejected response. A tied score counts as wrong in every accuracy. The search holds out whole conversations until at least 10\% of the pairs are held out (6,000 fit and 667 held-out pairs over 609 conversations), and the final arms retrain on all 6,667 pairs with seeds 0 to 4. Training inputs were truncated at 2,048 tokens, which affected 13 of the 13,334 training sequences (8 pairs). Scoring ran in bf16 on RTX 4090 cards and training on 96 GB cards with torch 2.14.0, transformers 5.17.0
and optuna 4.5.0.

\paragraph{Cross-fitting and intervals.}
The calibrated variance, the response and hidden-state probes and the prompt probe use five grouped folds, grouped by
the normalized first turn on HelpSteer2 and by the prompt on HelpSteer, with the ridge penalty chosen inside each
fold. The partial-centering weight is chosen on one of two prompt folds and applied to the other. The bootstrap
intervals of these analyses hold the fitted calibration, residuals or weights fixed and resample prompts, so they are
conditional on the fitted transformation. The two-way decomposition of the transplant scores refits in every resample.

\paragraph{Independent reference responses.}
The pool holds six responses from Qwen3-1.7B and six from Llama-3.2-3B-Instruct for each of the 378 HelpSteer prompts,
sampled at temperature 0.7 and top-p 0.9 with at most 2,048 new tokens under each model's chat template, and kept
without filtering (two responses stopped at the cap). The pool's responses are two to three times longer than the
rated ones (median 145 and 169 words against 60). Partial centering subtracts $\lambda$ times the prompt's pool mean,
with $\lambda$ from 0 to 1 in steps of 0.1 chosen by the selected set's mean helpfulness on the other fold. The chosen
value is 0 on both folds for four models and 0.1 on one fold for Tulu-3 and URM in that run. The recomputation behind Table~\ref{tab:primary} chooses 0.1 on one fold for Tulu-3 only. The shrinkage baseline subtracts
$\mu + B(\bar P_p - \mu)$, with $\bar P_p$ the prompt's pool mean, $\mu$ its mean over prompts and
$B = \tau^2 / (\tau^2 + s_w^2/12)$, where $s_w^2$ is the mean within-pool variance and $\tau^2$ the between-prompt
variance of the pool means net of $s_w^2/12$. $B$ is 0.96 to 0.98.

\paragraph{Leaderboard and filtering experiments.}
The five policies are Qwen3-0.6B, Qwen3-1.7B (the anchor), Qwen3-4B, Llama-3.2-1B-Instruct and Llama-3.2-3B-Instruct
as released, decoding greedily with at most 2,048 new tokens on the 356 single-turn validation prompts. The four
prompt mixes are the prompts above and at or below the median word count, and the two halves of a twenty-cluster
TF-IDF k-means over the first turns. Each of the twenty assignments gives every policy one mix drawn uniformly, and
an assignment that puts all five on the same mix is redrawn. The judge sees the prompt and the two responses labelled
A and B, is asked "Which response is more helpful to the user? Answer with the letter A or B only.", and its answer is the probability of A renormalized over the two letters, averaged over both orders. The filtering policies were trained with DPO ($\beta = 0.1$, two epochs, 32 pairs per step, 2,449 pairs per arm), as a full fine-tune judged by the 70B judge and as a low-rank variant (rank 16) judged by the 24B judge, with the learning rate chosen in each case on the random arm's held-out conversations.

\section{Supporting results}
\label{app:support}

\paragraph{Feature probes and variance decomposition.}
Residualizing each reward on the shown response's complexity and verbosity ratings changes the share by at most
0.02, and residualizing on all five ratings raises it by 0.01 to 0.10 on HelpSteer2 validation. The five ratings
account for 0.06 to 0.33 of each reward's between-prompt variance there (0.20 to 0.46 on HelpSteer) and for 0.31 to
0.42 of its within-prompt variance, so the residual share rises while most of the prompt-level variance stays
(Table~\ref{tab:components}). Four learners (ridge, kernel ridge, boosted trees, a multilayer perceptron) on two
sentence embeddings of the response with text features explain 0.07 to 0.35 of the reward variance cross-validated
by prompt, and across the four learners the residual keeps a between-prompt share of 0.26 to 0.55, at most 0.10 below the reward's own. From
the prompt side, a ridge probe from a sentence embedding of the prompt to the per-prompt mean reward, cross-fitted
by prompt, explains 0.07 to 0.31 of its variance for the six models on HelpSteer2. Prompt length and turn count
explain 0.01 to 0.24 of it. The same probe fitted in sample reaches 0.38 to 0.53 from the embedding and 0.69 to
0.79 from the last-token state of the Qwen3-8B base model on the three pairwise models, and cross-fitted the base-model states explain 0.11 to 0.40. These in-sample and cross-fitted estimates use our response sets and are not directly comparable with the
held-out, 32-response setting of \citet{hasanaliyev2026expected}. For three models, a cross-fitted probe from the model's own last-token state under the generic prompt
explains 0.40, 0.38 and 0.65 of the per-prompt mean real reward, and the residual keeps a share of 0.32, 0.11 and
0.23. In a crossed fit of the scores over own, donor and generic prompts with an indicator for the original prompt, the
match between a response and its own prompt is the largest term, 4 to 18 times the response effect, and which
wrong prompt is attached to a fixed response moves its score about as much as which response is scored.

\paragraph{Leaderboard, filtering and history checks.}
Across twenty assignments of five policies to four single-turn prompt mixes, the ranking by raw mean score differs
from the full-coverage ranking in 0 to 12 of the 20 assignments per reward model, and centering changes its agreement with a 70B judge
by at most 0.10 in Kendall's tau, with every interval including zero (Table~\ref{tab:leaderboard}). The judge agrees with the human helpfulness order on 0.78 [0.73, 0.83] of 252 untied pairs. Raw and centered filters select substantially different training sets, and the downstream evaluation does not
establish a difference between the resulting policies: the 30\% of a training pool kept by raw score and by centered score overlap at a
Jaccard index of 0.30, and preference training on either set from a 1.7B base, with a full fine-tune or a low-rank update, gives policies that win 0.50 to 0.51 of judged comparisons against the base, with every difference's interval including zero, and neither judge prefers one policy over the other (0.51 [0.49, 0.54] and 0.51 [0.49, 0.52]). For the low-rank arms, held-out implicit-reward accuracy is 0.66 to 0.69 against a chance level of 0.5, and 2 to 7 percent of responses are identical to the base's, while the two judges agree with themselves across presentation orders on 0.22 to 0.27 and 0.08 to 0.12 of these prompts, so the evaluation establishes neither a difference nor equivalence. Removing the history changes the scores: with the last user turn only, the final response's score moves by 0.57
to 1.08 standard deviations for all seventeen models scored, and the share under that input is 0.46 to 0.71,
above the single-turn share by 0.14 to 0.48 for sixteen of the seventeen (Table~\ref{tab:history}).

\paragraph{Human references and rating reliability.}
The HelpSteer2 training-split share rests on 10,160 prompts and is shown as context beside the validation value.
Each released HelpSteer2 rating averages three raters with a reliability of 0.73. A single rater's reliability
there is 0.48 by Spearman--Brown over all raters, 0.62 as one annotator against the others and 0.78 over the three
raters kept per response, and the three-rater mean implied by 0.62 is 0.83. The noise-corrected HelpSteer value is
0.55, 0.42 [0.32, 0.53], 0.34 and 0.32 under these four reliability values. HelpSteer2's released label is the rounded mean of the three
most agreeing of three to five annotations, so these reliabilities describe the raters and not the released label
exactly, and the correction is an approximate sensitivity analysis. Under an independent additive-noise model the
between-prompt variance component itself is not attenuated, so no corrected variance is reported. On the Reddit posts the between-post share of log vote counts is 0.49. After the post's exposure is regressed out it is 0.39.

\paragraph{Correlation with prompt-level human ratings.}
Across prompts, a model's mean score correlates with the mean human rating at 0.40 to 0.58 on HelpSteer and 0.08
to 0.24 on the Reddit posts. With model and human means taken on disjoint halves of each prompt's responses it is
0.22 to 0.31 on HelpSteer (the correlation between mean human ratings on the two response halves is 0.38 [0.30, 0.46]) and 0.00 to 0.09 on the Reddit posts
(humans 0.54). On HelpSteer2 the whole-prompt correlation is 0.36 to 0.55 for the six core models, because both means
rest on the same two responses, and the disjoint-half version is within 0.08 of zero (the same correlation for the human ratings is $-0.10$). Under the generic prompt the HelpSteer correlation falls to $-0.03$ to 0.15. Table~\ref{tab:concordance} reports how the per-prompt mean scores change under the generic prompt.

\paragraph{Sensitivity to prompt-set composition.}
We split the 356 single-turn HelpSteer2 validation prompts at the median prompt length. The mean raw reward of the
responses is lower on the longer half by 0.02 to 0.39 standard deviations for every released model, with the
interval below zero for 15 of 17, while the human helpfulness differs by $-0.05$ $[-0.19, 0.08]$. The model
difference exceeds the humans' with an interval above zero for 10 of 17. On all prompts, where the longer half is
largely multi-turn, the shift is 0.21 to 0.73. Under the same generic prompt, substantial score differences remain between the responses to the longer and
the shorter prompts. Two balanced topic splits move
the mean by under 0.15 (Table~\ref{tab:shift}).

\paragraph{Reference-pool size and shrinkage.}
On HelpSteer, centering each score on one, two or three of its siblings costs 0.19 to 0.30, 0.14 to 0.23 and 0.14
to 0.21 points of the kept set's mean helpfulness at the top 30\%, with every interval below zero, and the pooled
Spearman correlation with the raw rating falls by 0.14 to 0.22, then 0.09 to 0.16. The cost shrinks for every
model as the baseline grows. A $1/m$ extrapolation puts its limit at 0.10 to 0.17 points, which assumes that the
selection cost is linear in the baseline's variance. With the independent pool, a
shrinkage baseline that shrinks each prompt's pool mean toward the global mean by an empirical-Bayes weight of
0.96 to 0.98 costs 0.04 to 0.31 points (Table~\ref{tab:pool}).

\paragraph{Scoring robustness.}
Removing the end-of-sequence marker cuts the random-prompt margin by 1.4 to 3.0 standard deviations for four of the five core models that have one and by 0.57 for Tulu-3, and moves the shares by under 0.12 except URM's, which rises from 0.34 to 0.57. Removing the start marker moves margins by at most 0.12 and shares by at most 0.03
(Table~\ref{tab:templates}). Three instead of four random donors changes the match accuracy by under 0.01 for every
8B model but DPA (+0.011). With a fixed reply in place of the response, the original prompt wins against random donors at
chance (0.19 to 0.21) and against its three related tasks at chance for five models and above chance for DPA
(0.43) and Skywork-Llama-8B (0.47), against 0.94 to 0.97 with the original response, so for those two part of the
related-task margin remains when the response is fixed. A 24B instruction-tuned judge rates 0.90 of the 1,407 paraphrases
as preserving the task, and restricted to those the mean shift changes by at most 0.016 standard deviations. The 50
prompts without a valid paraphrase are longer than the kept ones (median 113 against 40 words) and half of them are multi-turn. The share without them is within 0.02 of the full value for 15 of 17 released models and within 0.05
for all. ArmoRM's combined output has a share of 0.81 on HelpSteer2 and 0.65 on HelpSteer against the
helpfulness head's 0.51 and 0.49, the same generic-prompt share, and a smaller margin (2.3 against 3.6 standard deviations). The paper scores the head. The longest input in any validation condition is 3,746 tokens against the
4,096-token cap, so no scored row was truncated.

\paragraph{Second-judge leaderboard check.}
A 24B judge of another family ranks the five policies as the 70B judge does (Kendall's tau 1.00; pairwise agreement
0.885 over 1,424 judged pairs) and gives the same reading of the leaderboard, with the centered ranking's tau
higher than the raw one's by $-0.00$ to 0.09.

\paragraph{Additional training conditions.}
At 0.6B, a Bradley--Terry arm trained on the rating margin has a standardized mismatch margin of 2.34, between the two main arms, with a share of 0.34, and a regression arm trained
on binarized ratings loses most of the margin (0.71) along with pair accuracy (0.650 against 0.686)
(Table~\ref{tab:arms}). At 1.7B the rating-margin arm reaches 2.60 standard deviations (across seeds 2.24 to 3.00)
with a share of 0.49 (0.43 to 0.59), between the two main arms on both means, while its seed ranges overlap both
main arms' on the margin and the Bradley--Terry arm's on the share. The binarized regression arm at 1.7B has a
held-out pair accuracy of 0.753 against 0.745 and 0.748 for the two main arms but has a standardized mismatch margin of 1.95
(1.79 to 2.03), 0.81 under the regression arm with every seed below every regression seed and 0.06 under the
Bradley--Terry arm, and its share of 0.13 (0.11 to 0.16) is the lowest of any 1.7B arm.

\section{Complete results and model comparison}
\label{app:tables}

\paragraph{Audit of the selection panels.}
Reference centering subtracts one offset per prompt. Over eighteen centerings of the 378 HelpSteer prompts for each
scorer (WildReward-8B, the five 1.7B true-mean seeds, and the six core models on a re-scoring of the pool on a
second GPU type), the within-prompt spread of the offset is at most $3.6 \times 10^{-15}$ and no within-prompt order
changes. That re-scoring reproduces the six centering effects of Table~\ref{tab:primary} within 0.005 points with
every interval below zero.
Candidates and reference responses are sampled separately. Of the 4,536 HelpSteer references, 25 repeat a candidate's text verbatim, and 173 of
the 3,246 IFEval references do, short answers sampled twice rather than shared responses. Of the 2,268 Phi-3.5-mini references, 7 repeat a candidate's text and 7 stop at the length cap. No RoSE reference equals a
system summary. The partial-centering weight runs from 0 (raw) to 1 (full centering) in steps of 0.1, chosen at 30
percent retention on one fold and applied to the other. The chosen weight is 0 for every model and fold except
Tulu-3 on one fold (0.1). On RoSE every scorer input is the fixed instruction followed by the full released article
(no exception in 13,984 inputs), no row is truncated (the longest input is 3,313 tokens), the released ``gold'' system
appears only in the twelve-system test extension, and one validation article (eight system rows) contains its own
reference summary inside the released article text, with none in the test panel. On IFEval the checker is pinned at one commit, 218 prompts pass the strict check
on all six candidates and 54 fail on all six (244 and 38 under the loose check), all kept. Of the 3,246 generated
candidates, 172 reached the length cap (142 from Qwen3-1.7B and 30 from Llama-3.2-3B-Instruct), and DPA truncated
four rows while the other five models truncated none. The audit file will be released with the protocol, whose implementation regenerates the primary contrasts and the
selected identities of Table~\ref{tab:primary} byte-identically in a clean environment and passes six tests. They check that one offset
per prompt keeps every within-prompt order, that an increasing global transform keeps the selected set, that the
tie rule is deterministic and label-free, that the kept count is exact, that the prefix oracle equals brute force
over all allocations. The sixth test checks that the exchange identity reconstructs the utility difference.

\paragraph{Data and model terms.}
HelpSteer2 and HelpSteer are used under CC-BY-4.0, RewardBench and its second version under ODC-By, RoSE and IFEval
under their released licences, and the Stanford Human Preferences set under the terms its card states for Reddit
content, for research only and not redistributed. Model weights are used under their own licences.

\begin{figure*}[htbp]
\centering
\includegraphics[width=\textwidth]{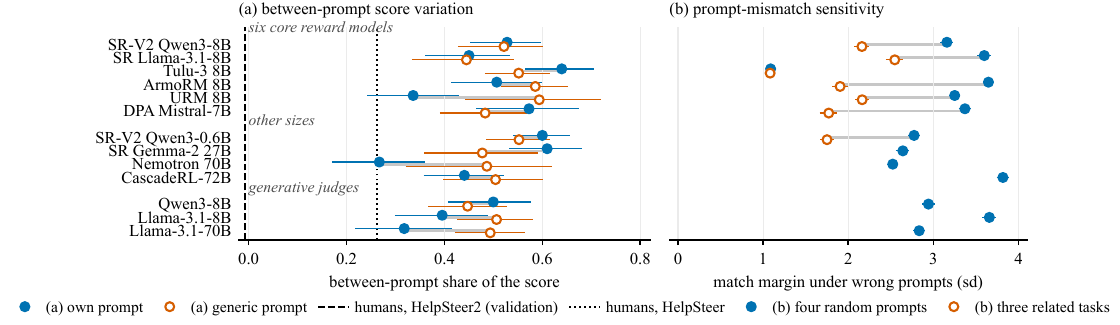}
\caption{Between-prompt variation and mismatch sensitivity on HelpSteer2 validation for thirteen of the eighteen released scorers. (a) The share under the original prompt and under one generic prompt is drawn against the human validation references, and the HelpSteer reference comes from a different response population. (b) The mean score loss, in original-condition standard deviations, under four random wrong prompts and, for seven models, under three related tasks written by an instruction model. Bars are 95\% prompt-bootstrap intervals. The size ladders are in Figure~\ref{fig:share}.}
\label{fig:overview}
\end{figure*}\begin{figure*}[htbp]
\centering
\includegraphics[width=\textwidth]{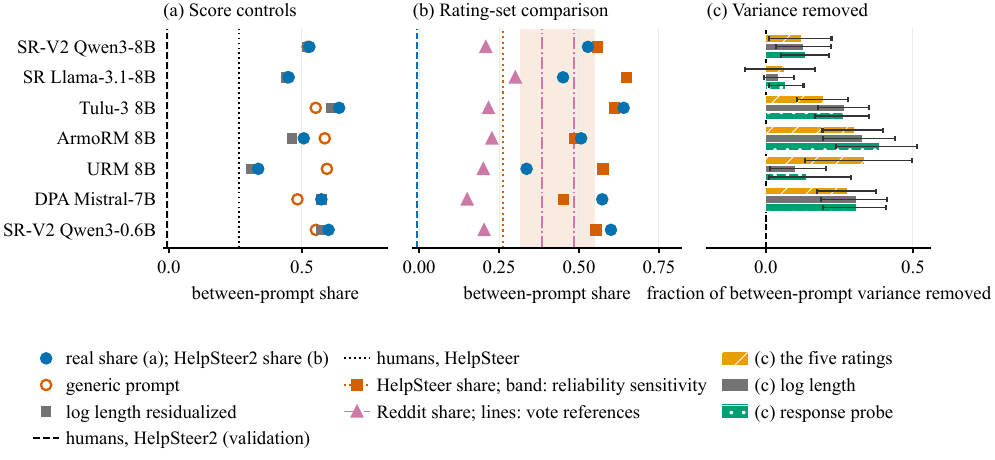}
\caption{Controls and human references for seven reward models. (a) HelpSteer2 variance shares under the original prompt, a generic prompt and log-length adjustment. (b) Shares on three rating sets with each set's human reference; the band is HelpSteer's rating-reliability sensitivity analysis, and the Reddit references use raw and exposure-adjusted log votes. (c) Fraction of HelpSteer2 between-prompt variance removed by each adjustment for the six core models. Bars are 95 percent prompt-bootstrap intervals.}
\label{fig:controls}
\end{figure*}\begin{figure*}[htbp]
\centering
\includegraphics[width=\textwidth]{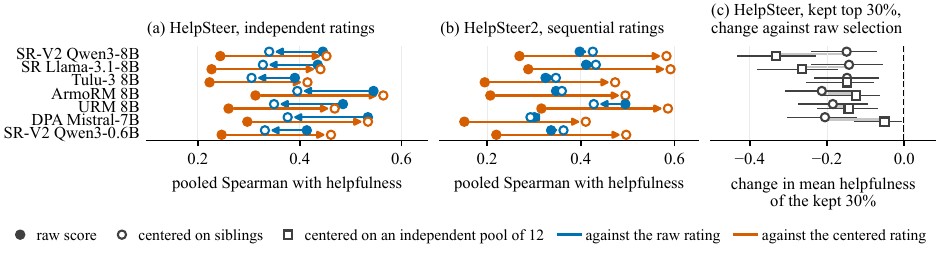}
\caption{Centering and global selection. (a, b) Each arrow runs from the raw score (filled) to the sibling-centered score (open) and shows the pooled Spearman correlation with the raw helpfulness rating and with the rating centered the same way, on HelpSteer (independent ratings) and HelpSteer2 validation (sequential ratings). (c) The change in the mean helpfulness (0 to 4) of the 30\% of HelpSteer responses kept, against raw selection, under sibling centering and under centering on an independent pool of twelve generated responses. Panels (a, b) show seven models and (c) the six core models. Bars in (c) are paired 95\% prompt-bootstrap intervals. The values are in Tables~\ref{tab:centering} and \ref{tab:pool}. Panel (c) draws a separate run of the centering sweep, whose twelve-reference effects differ from Table~\ref{tab:primary} by at most 0.016.}
\label{fig:centering}
\end{figure*}\begin{figure*}[htbp]
\centering
\includegraphics[width=\textwidth]{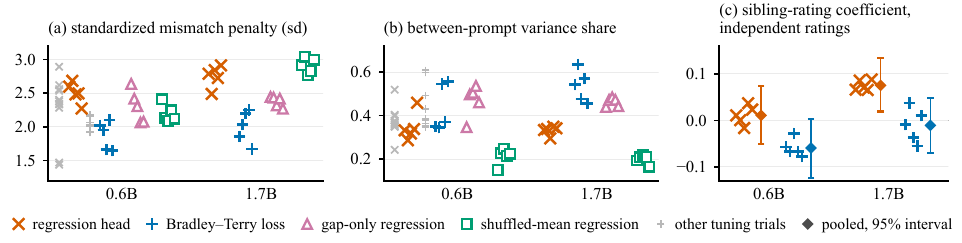}
\caption{Matched training in Qwen3 (0.6B and 1.7B, five seeds per arm). (a) The standardized mismatch penalty under four random wrong prompts. (b) The between-prompt share on HelpSteer2 validation. (c) The sibling-rating coefficient on HelpSteer. Panels (a, b) show four training conditions and (c) the two main objectives. Points are seeds, and the diamonds and bars in (c) are pooled estimates with 95\% intervals. Grey marks are the other 0.6B tuning trials. Table~\ref{tab:arms} reports all six training conditions and the two configuration swaps.}
\label{fig:matched}
\end{figure*}

\begin{table*}[htbp]
\centering
\footnotesize
\setlength{\tabcolsep}{1.5pt}
\begin{tabular}{lrrrrrrrr}
\toprule
Reward model & Share & Single-turn & Generic & Length removed & Prompt only & Margin & Accuracy & Related task \\
\midrule
Humans, HelpSteer2 train & 0.02 & 0.01 & & & & & & \\
Humans, HelpSteer2 validation & -0.01 & -0.07 & & & & & & \\
Humans, HelpSteer validation & 0.26 & 0.26 & & & & & & \\
\midrule
SR-V2-Qwen3-8B & 0.53 & 0.43 & 0.52 & 0.52 & 0.26 & 3.16 & 0.99 & 2.16 \\
SR-Llama-3.1-8B-v0.2$^\dagger$ & 0.45 & 0.36 & 0.44 & 0.44 & 0.34 & 3.60 & 0.98 & 2.55 \\
Tulu-3-8B-RM & 0.64 & 0.45 & 0.55 & 0.61 & 0.40 & 1.09 & 0.88 & 1.08 \\
ArmoRM-Llama3-8B-v0.1 & 0.51 & 0.43 & 0.59 & 0.46 & -0.01 & 3.65 & 0.99 & 1.90 \\
URM-LLaMa-3.1-8B & 0.34 & 0.28 & 0.59 & 0.31 & 0.18 & 3.25 & 0.99 & 2.16 \\
DPA-Mistral-7B-v1 & 0.57 & 0.36 & 0.48 & 0.57 & -0.30 & 3.37 & 0.94 & 1.77 \\
SR-V2-Qwen3-0.6B & 0.60 & 0.55 & 0.55 & 0.57 & 0.11 & 2.77 & 0.99 & 1.75 \\
\bottomrule
\end{tabular}
\caption{The HelpSteer2 validation audit for the six core reward models and the 0.6B checkpoint, after three human reference rows. The share columns are between-prompt variance shares on all prompts, on single-turn prompts, under one generic prompt and with log length residualized. The prompt-only column is the correlation of a fixed reply's score with the prompt means. The margin and accuracy columns use four random wrong prompts and the last column three related tasks, in original-condition standard deviations. The dagger marks the model whose preference data contain 94 of the 511 validation first turns, and the 27B model uses the same preference set. Table~\ref{tab:conditions-full} gives intervals.}
\label{tab:conditions}
\end{table*}

\begin{table}[htbp]
\centering
\small
\begin{tabular}{p{0.55\columnwidth}p{0.33\columnwidth}}
\toprule
Operation & Effect of a shared prompt shift \\
\midrule
Pairwise comparison within a prompt & none \\
Best-of-$n$ within a prompt & none \\
Two policies compared on the same prompts with one scorer & none \\
Policy means ranked on identical prompts & cancels \\
Keep the top fraction across prompts & changes the kept set \\
Accept above one global threshold & changes acceptance and quality \\
\bottomrule
\end{tabular}
\caption{Effect of adding the same constant to every response score of a prompt on score-based operations.}
\label{tab:operations}
\end{table}

\begin{table*}[htbp]
\centering
\footnotesize
\resizebox{\textwidth}{!}{
\begin{tabular}{llr}
\toprule
Panel, criterion & Contrast at 30\% retention & Effect [95\% interval] \\
\midrule
HelpSteer, helpfulness & true minus permuted means & +0.078 [+0.039, +0.136] \\
HelpSteer, helpfulness & true minus no mean term & +0.060 [+0.028, +0.103] \\
HelpSteer, helpfulness & true means minus Bradley--Terry & +0.180 [+0.124, +0.244] \\
HelpSteer, helpfulness & centered minus raw, Skywork-Reward-V2-Qwen3-8B & -0.333 [-0.430, -0.229] \\
HelpSteer, helpfulness & centered minus raw, Skywork-Reward-Llama-3.1-8B-v0.2 & -0.264 [-0.379, -0.174] \\
HelpSteer, helpfulness & centered minus raw, Llama-3.1-Tulu-3-8B-RM & -0.148 [-0.253, -0.079] \\
HelpSteer, helpfulness & centered minus raw, ArmoRM-Llama3-8B-v0.1 & -0.128 [-0.216, -0.068] \\
\bottomrule
\end{tabular}}
\caption{Primary contrasts at 30 percent retention with paired 95 percent prompt-bootstrap intervals, in each
criterion's units. The training contrasts are seed means over five matched seed pairs of the pair-trained 1.7B arms.}
\label{tab:primary}
\end{table*}

\begin{figure*}[htbp]
\centering
\includegraphics[width=\textwidth]{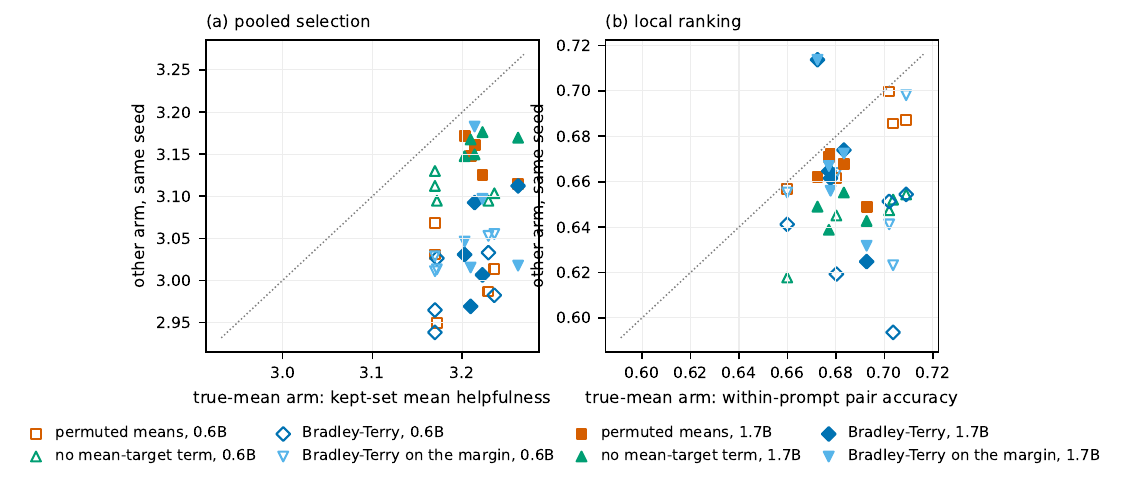}
\caption{Pair-trained matched arms on HelpSteer validation, each seed plotted against the true-mean arm of the
same seed: (a) kept-set mean helpfulness at 30 percent retention, (b) within-prompt pair accuracy. Hollow markers
are 0.6B, filled 1.7B. The dotted line is equality. Values are in Table~\ref{tab:armslocal}.}
\label{fig:supervision}
\end{figure*}

\begin{table*}[htbp]
\centering
\footnotesize
\resizebox{\textwidth}{!}{
\begin{tabular}{llrrr}
\toprule
Size & Arm & Kept-set mean helpfulness [seed range] & Within-prompt accuracy [seed range] & Seeds \\
\midrule
1.7B & true means (pointwise regression) & 3.222 [3.203, 3.262] & 0.681 [0.672, 0.693] & 5 \\
1.7B & permuted means & 3.144 [3.115, 3.172] & 0.664 [0.649, 0.672] & 5 \\
1.7B & no mean-target term & 3.162 [3.148, 3.176] & 0.650 [0.639, 0.665] & 5 \\
1.7B & Bradley--Terry & 3.042 [2.969, 3.112] & 0.668 [0.625, 0.714] & 5 \\
1.7B & Bradley--Terry on the rating margin & 3.072 [3.015, 3.183] & 0.668 [0.632, 0.714] & 5 \\
1.7B & regression on binarized ratings & 3.054 [3.033, 3.068] & 0.629 [0.623, 0.636] & 5 \\
0.6B & true means (pointwise regression) & 3.195 [3.170, 3.236] & 0.691 [0.660, 0.709] & 5 \\
0.6B & permuted means & 3.010 [2.949, 3.068] & 0.678 [0.657, 0.700] & 5 \\
0.6B & no mean-target term & 3.107 [3.095, 3.130] & 0.643 [0.618, 0.654] & 5 \\
0.6B & Bradley--Terry & 2.989 [2.938, 3.033] & 0.632 [0.594, 0.654] & 5 \\
0.6B & Bradley--Terry on the rating margin & 3.032 [3.011, 3.055] & 0.656 [0.623, 0.698] & 5 \\
0.6B & regression on binarized ratings & 2.956 [2.916, 3.042] & 0.611 [0.515, 0.665] & 5 \\
\bottomrule
\end{tabular}}
\caption{Pair-trained matched arms on HelpSteer validation: kept-set mean helpfulness at 30 percent retention and
within-prompt pair accuracy, seed means with the seed range.}
\label{tab:armslocal}
\end{table*}

\begin{table*}[htbp]
\centering
\footnotesize
\resizebox{\textwidth}{!}{
\begin{tabular}{lrrrr}
\toprule
Arm pair (1.7B) & Pair-order disagreement & Pair accuracy, A minus B & Best-of-four, A minus B & Pooled 30\%, A minus B \\
\midrule
true means vs permuted means & 0.160 & +0.016 [+0.001, +0.033] & +0.021 [-0.012, +0.053] & +0.078 \\
true means vs no mean term & 0.160 & +0.030 [+0.015, +0.048] & +0.043 [+0.007, +0.082] & +0.060 \\
true means vs Bradley--Terry & 0.214 & +0.013 [-0.008, +0.033] & +0.016 [-0.021, +0.055] & +0.180 \\
permuted means vs no mean term & 0.165 & +0.014 [-0.002, +0.030] & +0.022 [-0.011, +0.057] & -- \\
permuted means vs Bradley--Terry & 0.216 & -0.003 [-0.024, +0.018] & -0.005 [-0.045, +0.032] & -- \\
no mean term vs Bradley--Terry & 0.202 & -0.018 [-0.037, +0.001] & -0.027 [-0.068, +0.016] & -- \\
\bottomrule
\end{tabular}}
\caption{Local controls between the pair-trained 1.7B arms on HelpSteer validation, seed means with paired 95
percent prompt-bootstrap intervals: the fraction of within-prompt pairs ordered differently, the pair-accuracy
difference, the difference in each prompt's best-of-four rating, and the pooled difference at 30 percent retention.
Retrains of one configuration disagree on 0.15 to 0.19 of pairs.}
\label{tab:localcontrols}
\end{table*}

\begin{table*}[htbp]
\centering
\footnotesize
\resizebox{\textwidth}{!}{
\begin{tabular}{lrrr}
\toprule
Size, condition or contrast & Test kept mean [seed range] & Test pair accuracy & Validation kept mean [seed range] \\
\midrule
1.7B, true means & 3.345 [3.330, 3.357] & 0.753 & 3.298 [3.280, 3.317] \\
1.7B, permuted means & 3.234 [3.213, 3.252] & 0.747 & 3.216 [3.198, 3.242] \\
1.7B, no mean term & 3.226 [3.220, 3.238] & 0.761 & 3.187 [3.172, 3.200] \\
1.7B, true minus omitted & +0.118 [+0.072, +0.160], 5 of 5 seeds & & +0.112 [+0.032, +0.194], 5 of 5 seeds \\
1.7B, true minus permuted & +0.111 [+0.062, +0.157], 5 of 5 seeds & & +0.082 [+0.010, +0.162], 5 of 5 seeds \\
\midrule
0.6B, true means & 3.297 [3.276, 3.317] & 0.734 & 3.270 [3.251, 3.288] \\
0.6B, permuted means & 3.175 [3.122, 3.221] & 0.737 & 3.146 [3.119, 3.181] \\
0.6B, no mean term & 3.202 [3.183, 3.222] & 0.742 & 3.167 [3.139, 3.185] \\
0.6B, true minus omitted & +0.095 [+0.050, +0.139], 5 of 5 seeds & & +0.103 [+0.042, +0.185], 5 of 5 seeds \\
0.6B, true minus permuted & +0.122 [+0.067, +0.180], 5 of 5 seeds & & +0.124 [+0.053, +0.201], 5 of 5 seeds \\
\midrule
random selection (200 draws) & 2.850 & & 2.827 \\
label-sort bound & 3.808 & & 3.733 \\
\bottomrule
\end{tabular}}
\caption{The three conditions trained on HelpSteer's training split at two sizes, five seeds each: kept mean
helpfulness at 30 percent retention on the locked test groups (1,123 prompts, 4,492 responses) and on HelpSteer
validation, with the seed range, within-prompt pair accuracy on the test groups, and the two contrasts with
Bonferroni-adjusted 95 percent intervals from a crossed prompt-and-seed bootstrap.}
\label{tab:ea2}
\end{table*}

\begin{table*}[htbp]
\centering
\footnotesize
\resizebox{\textwidth}{!}{
\begin{tabular}{lrrr}
\toprule
Arm (1.7B, retrained set) & Raw kept mean [seed range] & Centered kept mean [seed range] & Change \\
\midrule
true means & 3.222 [3.203, 3.262] & 3.081 [3.040, 3.110] & -0.141 \\
no mean term (retrained) & 3.180 [3.161, 3.198] & 3.021 [3.007, 3.031] & -0.159 \\
permuted means (retrained) & 3.119 [3.101, 3.132] & 3.051 [3.015, 3.073] & -0.067 \\
\midrule
true minus permuted means & +0.103 [+0.058, +0.158], 5 of 5 seeds & +0.030 [-0.004, +0.081], 4 of 5 seeds & \\
true minus no mean term & +0.042 [+0.011, +0.092], 5 of 5 seeds & +0.060 [+0.018, +0.103], 5 of 5 seeds & \\
\bottomrule
\end{tabular}}
\caption{Pair-trained 1.7B arms under raw and reference-centered selection at 30 percent retention on HelpSteer
validation: the surviving true-mean arm and retrained permuted-mean and omitted-mean arms, five seeds each, scored
on the twelve-reference pool. Seed means with the seed range, and the two contrasts with paired 95 percent
prompt-bootstrap intervals. The retrained arms are a separate set from those of Table~\ref{tab:primary}.}
\label{tab:armscentered}
\end{table*}

\begin{table*}[htbp]
\centering
\footnotesize
\resizebox{\textwidth}{!}{
\begin{tabular}{llrr}
\toprule
Model & Predictor (fitted on one fold, read on the other) & Kept mean & Minus raw [95\% interval] \\
\midrule
SR-V2-Qwen3-8B & raw-only affine & 3.282 & +0.000 [+0.000, +0.000] \\
SR-V2-Qwen3-8B & centered-only & 2.936 & -0.346 [-0.445, -0.227] \\
SR-V2-Qwen3-8B & reference-aware & 3.282 & +0.000 [-0.020, +0.037] \\
SR-V2-Qwen3-8B & prompt-only & 3.053 & -0.229 [-0.302, -0.119] \\
SR-V2-Qwen3-8B & logistic $P(H \geq 2 \mid r, m)$ & 3.260 & -0.022 [-0.042, +0.033] \\
SR-V2-Qwen3-8B & residual head $u_0 + g(x)$ (bge-small) & 3.238 & -0.044 [-0.073, +0.024] \\
SR-Llama-3.1-8B & raw-only affine & 3.236 & +0.000 [+0.000, +0.000] \\
\bottomrule
\end{tabular}}
\caption{Calibrated predictors against raw scoring on HelpSteer validation, fitted on one prompt fold and evaluated
on the other. The reference-aware form is $u = a + b_W (r - m) + b_B m$ with $m$ the mean of the twelve references.
Intervals are conditional on the fits.}
\label{tab:calibration}
\end{table*}

\begin{table*}[htbp]
\centering
\footnotesize
\resizebox{\textwidth}{!}{
\begin{tabular}{llrrrr}
\toprule
Model & References & Kept mean & Minus raw & Minus all twelve & Exchanged against all twelve \\
\midrule
SR-V2-Qwen3-8B & all12 & 2.941 & -0.333 [-0.430, -0.229] & +0.000 [+0.000, +0.000] & 0.00 \\
SR-V2-Qwen3-8B & qwen6 & 2.914 & -0.359 [-0.454, -0.249] & -0.026 [-0.075, +0.026] & 0.13 \\
SR-V2-Qwen3-8B & llama6 & 2.985 & -0.288 [-0.403, -0.196] & +0.044 [-0.020, +0.090] & 0.17 \\
SR-V2-Qwen3-8B & mix3+3 draw 0 & 2.954 & -0.319 [-0.425, -0.227] & +0.013 [-0.029, +0.040] & 0.04 \\
SR-V2-Qwen3-8B & mix3+3 draw 1 & 2.958 & -0.315 [-0.423, -0.225] & +0.018 [-0.029, +0.044] & 0.05 \\
SR-V2-Qwen3-8B & mix3+3 draw 2 & 2.963 & -0.311 [-0.412, -0.205] & +0.022 [-0.011, +0.055] & 0.04 \\
SR-V2-Qwen3-8B & mix3+3 draw 3 & 2.941 & -0.333 [-0.432, -0.227] & +0.000 [-0.037, +0.044] & 0.04 \\
\bottomrule
\end{tabular}}
\caption{Reference composition on HelpSteer validation at 30 percent retention: six Qwen3-1.7B references, six
Llama-3.2-3B-Instruct references, five preselected three-plus-three subsets (averaged) and all twelve, against raw
scoring and against the twelve-reference selection, with the fraction of the kept set exchanged.}
\label{tab:composition}
\end{table*}

\begin{table*}[htbp]
\centering
\footnotesize
\begin{tabular}{llrrr}
\toprule
Model & References & Kept mean & Minus raw [95\% interval] & Minus the original twelve \\
\midrule
SR-V2-Qwen3-8B & twelve (original) & 2.941 & -0.333 [-0.430, -0.229] &  \\
SR-V2-Qwen3-8B & eighteen & 2.976 & -0.297 [-0.388, -0.203] & +0.035 [-0.007, +0.084] \\
SR-V2-Qwen3-8B & twelve, Phi replaces Qwen & 2.985 & -0.288 [-0.379, -0.196] & +0.044 [-0.013, +0.112] \\
SR-V2-Qwen3-8B & twelve, Phi replaces Llama & 3.000 & -0.273 [-0.372, -0.187] & +0.059 [+0.004, +0.106] \\
SR-V2-Qwen3-8B & six (Phi only) & 3.018 & -0.256 [-0.355, -0.165] & +0.077 [-0.004, +0.150] \\
\midrule
SR-Llama-3.1-8B & twelve (original) & 2.976 & -0.264 [-0.374, -0.176] &  \\
SR-Llama-3.1-8B & eighteen & 2.993 & -0.247 [-0.337, -0.148] & +0.018 [-0.013, +0.075] \\
\bottomrule
\end{tabular}
\caption{Reference-pool composition with six Phi-3.5-mini responses per HelpSteer validation prompt, at 30 percent
retention: kept mean helpfulness, its difference from raw scoring and from centering on the original twelve, with
paired 95 percent prompt-bootstrap intervals. The original pool was rescored alongside the new references, and its
effects differ from Table~\ref{tab:primary} by at most 0.005 points.}
\label{tab:thirdfamily}
\end{table*}

\begin{table*}[htbp]
\centering
\footnotesize
(a) Between-prompt variance shares\\[2pt]
\begin{tabular}{lrrrr}
\toprule
Judge & HelpSteer2, original & HelpSteer2, generic & HelpSteer, original & HelpSteer, generic \\
\midrule
Claude Sonnet 5.5 & 0.46 [0.37, 0.55] & 0.45 [0.37, 0.52] & 0.60 [0.54, 0.66] & 0.61 [0.56, 0.66] \\
Claude Haiku 5.5 & 0.50 [0.42, 0.58] & 0.54 [0.47, 0.60] & 0.55 [0.49, 0.61] & 0.61 [0.56, 0.65] \\
\bottomrule
\end{tabular}\\[8pt]
(b) HelpSteer selection at 30 percent retention\\[2pt]
\begin{tabular}{lrrrrr}
\toprule
Judge & Raw selected mean & Centered minus raw & Pair accuracy & Tied fraction & Untied accuracy \\
\midrule
Claude Sonnet 5.5 & 3.231 & -0.098 [-0.222, -0.069] & 0.481 & 0.36 & 0.750 \\
Claude Haiku 5.5 & 3.264 & -0.215 [-0.308, -0.131] & 0.551 & 0.27 & 0.758 \\
\bottomrule
\end{tabular}
\caption{Closed judges on HelpSteer2 and HelpSteer validation, scored by the mean of ten sampled rating digits. (a)
Between-prompt variance shares under original and generic prompts. (b) HelpSteer selection at 30 percent retention:
raw mean helpfulness, the centering effect, and within-prompt accuracy. Score ties count as errors, and untied
accuracy excludes them. Brackets show 95 percent prompt-bootstrap intervals.}
\label{tab:closed}
\end{table*}

\begin{table*}[htbp]
\centering
\footnotesize
\resizebox{\textwidth}{!}{
\begin{tabular}{llllrr}
\toprule
Prompts & Fold & Choice rule on the calibration fold & Chosen scorer & Calibration value & Held-out kept mean \\
\midrule
all prompts & 0 & within acc & ArmoRM-8B & 0.758 & 3.326 \\
all prompts & 0 & utility30 & ArmoRM-8B & 3.401 & 3.326 \\
all prompts & 0 & best of four & ArmoRM-8B & 3.238 & 3.326 \\
all prompts & 1 & within acc & DPA-Mistral-7B & 0.753 & 3.322 \\
all prompts & 1 & utility30 & ArmoRM-8B & 3.326 & 3.401 \\
all prompts & 1 & best of four & DPA-Mistral-7B & 3.206 & 3.322 \\
eligible prompts only & 0 & within acc & ArmoRM-8B & 0.758 & 3.258 \\
eligible prompts only & 0 & utility30 & ArmoRM-8B & 3.277 & 3.258 \\
eligible prompts only & 0 & best of four & ArmoRM-8B & 3.226 & 3.258 \\
eligible prompts only & 1 & within acc & DPA-Mistral-7B & 0.753 & 3.194 \\
eligible prompts only & 1 & utility30 & DPA-Mistral-7B & 3.278 & 3.194 \\
eligible prompts only & 1 & best of four & DPA-Mistral-7B & 3.210 & 3.194 \\
\bottomrule
\end{tabular}}
\caption{Held-out model choice over nine scorers on HelpSteer validation: the scorer chosen on the calibration fold
by within-prompt accuracy, by kept quality at 30 percent and by its best-of-four rating, with each choice's held-out
kept mean. Eligible prompts exclude the 57 whose four responses share one rating.}
\label{tab:modelchoice}
\end{table*}

\begin{table*}[htbp]
\centering
\footnotesize
\resizebox{\textwidth}{!}{
\begin{tabular}{lllrrrrr}
\toprule
Model & Source fold & Target & Coverage, raw quantile & Kept mean & Coverage, risk threshold & Kept below-2 rate & Minus source rate \\
\midrule
SR-V2-Qwen3-8B & 0 & HelpSteer other fold & 0.257 & 3.216 & 0.257 & 0.026 & +0.021 \\
SR-V2-Qwen3-8B & 0 & HelpSteer2 validation & 0.460 & 3.342 & 0.460 & 0.040 & +0.035 \\
SR-V2-Qwen3-8B & 0 & HelpSteer source fold & 0.300 & 3.330 & 0.300 & 0.004 & +0.000 \\
SR-V2-Qwen3-8B & 1 & HelpSteer other fold & 0.329 & 3.309 & 0.329 & 0.004 & -0.018 \\
SR-V2-Qwen3-8B & 1 & HelpSteer2 validation & 0.488 & 3.320 & 0.488 & 0.049 & +0.027 \\
SR-V2-Qwen3-8B & 1 & HelpSteer source fold & 0.300 & 3.233 & 0.300 & 0.022 & +0.000 \\
SR-Llama-3.1-8B & 0 & HelpSteer other fold & 0.302 & 3.171 & 0.302 & 0.031 & +0.022 \\
\bottomrule
\end{tabular}}
\caption{Threshold transport. A raw-score quantile fixed at 30 percent coverage on one HelpSteer fold, and a
low-helpfulness risk threshold fitted on the same fold, applied to the other fold and to HelpSteer2 validation:
achieved coverage, kept mean helpfulness, and the kept set's below-2 rate against its source value.}
\label{tab:transport}
\end{table*}

\begin{figure*}[htbp]
\centering
\includegraphics[width=0.96\textwidth]{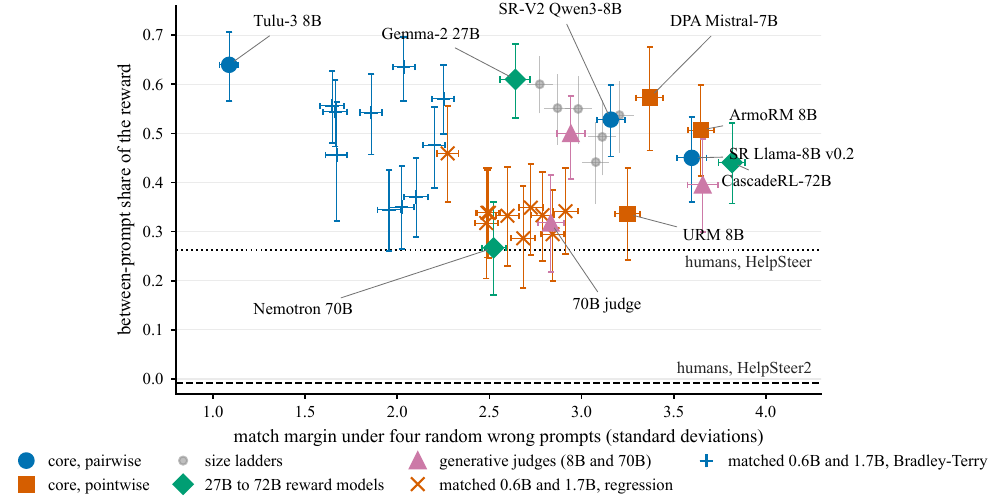}
\caption{All eighteen released scorers and the five seeds of each main matched arm, placed by their HelpSteer2 validation between-prompt share and their mismatch penalty under four random wrong prompts (original-condition standard deviations). The lines are the HelpSteer2 and HelpSteer validation references, and the HelpSteer reference comes from a different response population. Bars are 95\% prompt-bootstrap intervals. The other matched training conditions are in Table~\ref{tab:arms}.}
\label{fig:share}
\end{figure*}

\begin{table*}[htbp]
\centering
\footnotesize
\resizebox{\textwidth}{!}{
\begin{tabular}{lrrrrr}
\toprule
Attribute & HelpSteer2 train & HelpSteer2 validation & HelpSteer train & HelpSteer2 single-rater reliability & HelpSteer2 same-prompt correlation \\
\midrule
Helpfulness & \ci{0.02}{0.00}{0.04} & \ci{-0.01}{-0.11}{0.10} & \ci{0.24}{0.22}{0.25} & \ci{0.62}{0.61}{0.63} & \ci{-0.08}{-0.09}{-0.06} \\
Correctness & \ci{0.07}{0.05}{0.09} & \ci{0.00}{-0.10}{0.11} & \ci{0.29}{0.28}{0.31} & \ci{0.63}{0.62}{0.64} & \ci{-0.01}{-0.03}{0.00} \\
Coherence & \ci{0.08}{0.06}{0.11} & \ci{0.12}{-0.01}{0.24} & \ci{0.30}{0.29}{0.31} & \ci{0.26}{0.25}{0.28} & \ci{0.06}{0.04}{0.07} \\
Complexity & \ci{0.64}{0.62}{0.66} & \ci{0.63}{0.56}{0.70} & \ci{0.68}{0.67}{0.69} & \ci{0.31}{0.30}{0.32} & \ci{0.65}{0.64}{0.66} \\
Verbosity & \ci{0.00}{-0.02}{0.03} & \ci{0.04}{-0.07}{0.15} & \ci{0.46}{0.45}{0.48} & \ci{0.44}{0.43}{0.45} & \ci{0.02}{0.00}{0.04} \\
\bottomrule
\end{tabular}}
\caption{Human between-prompt shares by rating attribute and split on HelpSteer2 (train and validation) and on the HelpSteer training split, with HelpSteer2's single-rater reliability (one annotator against the others) and the correlation between the two responses' ratings for the same HelpSteer2 prompt. HelpSteer2 intervals use a separate bootstrap run from Table~\ref{tab:refs}, and one rounded endpoint differs.}
\label{tab:floor}
\end{table*}

Unless stated otherwise, brackets are 95\% percentile bootstrap intervals over prompts. Seed ranges are labelled where they are used. On HelpSteer2 the resampled unit is the conversation group (519 groups over 511 first turns, eight of which open two groups), and resampling the 511 first-turn clusters instead moves interval ends by at most 0.01. In the tables, "generic" is the generic-prompt
condition (one fixed instruction in place of every prompt), "prompt only" is the Spearman correlation across prompts
between a generic reply's reward and the mean real reward, margins are in standard deviations of the original-condition
rewards, and accuracy is the share of responses whose own prompt scores above every wrong prompt (chance 0.2 with
four, 0.25 with three). Figure~\ref{fig:share}
places every scored model on the two main estimates. Across the scalar reward models of that figure, the Spearman
correlation of the share with the RewardBench 2 Ties score is $-0.61$ $[-0.96, -0.02]$ over fourteen models, and
with RewardBench 1 and 2 it is $-0.46$ and $-0.47$ over fifteen, with intervals that include or touch zero. The
models include two size ladders and are not independent, so these correlations are descriptive and should be read with caution.

\begin{table*}[htbp]
\centering
\footnotesize
\resizebox{\textwidth}{!}{
\begin{tabular}{lrrrrrrrr}
\toprule
Reward model & Share & Single-turn & Generic & Length removed & Prompt only & Margin & Accuracy & Related task \\
\midrule
Humans, HelpSteer2 train & \ci{0.02}{0.00}{0.04} & \ci{0.01}{-0.02}{0.04} & & & & & & \\
Humans, HelpSteer2 validation & \ci{-0.01}{-0.11}{0.10} & \ci{-0.07}{-0.19}{0.04} & & & & & & \\
Humans, HelpSteer validation & \ci{0.26}{0.20}{0.33} & \ci{0.26}{0.20}{0.33} & & & & & & \\
\midrule
SR-V2-Qwen3-8B & \ci{0.53}{0.45}{0.60} & \ci{0.43}{0.33}{0.52} & \ci{0.52}{0.43}{0.60} & \ci{0.52}{0.44}{0.59} & \ci{0.26}{0.17}{0.34} & \ci{3.16}{3.08}{3.23} & \ci{0.99}{0.98}{0.99} & \ci{2.16}{2.07}{2.25} \\
SR-Llama-3.1-8B-v0.2$^\dagger$ & \ci{0.45}{0.36}{0.53} & \ci{0.36}{0.25}{0.46} & \ci{0.44}{0.33}{0.54} & \ci{0.44}{0.35}{0.53} & \ci{0.34}{0.26}{0.41} & \ci{3.60}{3.52}{3.68} & \ci{0.98}{0.98}{0.99} & \ci{2.55}{2.44}{2.65} \\
Tulu-3-8B-RM & \ci{0.64}{0.57}{0.71} & \ci{0.45}{0.34}{0.56} & \ci{0.55}{0.48}{0.61} & \ci{0.61}{0.53}{0.68} & \ci{0.40}{0.32}{0.48} & \ci{1.09}{1.04}{1.14} & \ci{0.88}{0.86}{0.91} & \ci{1.08}{1.04}{1.12} \\
ArmoRM-Llama3-8B-v0.1 & \ci{0.51}{0.41}{0.60} & \ci{0.43}{0.32}{0.54} & \ci{0.59}{0.52}{0.65} & \ci{0.46}{0.38}{0.55} & \ci{-0.01}{-0.10}{0.07} & \ci{3.65}{3.58}{3.72} & \ci{0.99}{0.98}{0.99} & \ci{1.90}{1.81}{2.00} \\
URM-LLaMa-3.1-8B & \ci{0.34}{0.24}{0.43} & \ci{0.28}{0.17}{0.39} & \ci{0.59}{0.44}{0.72} & \ci{0.31}{0.21}{0.41} & \ci{0.18}{0.10}{0.26} & \ci{3.25}{3.18}{3.32} & \ci{0.99}{0.98}{0.99} & \ci{2.16}{2.08}{2.24} \\
DPA-Mistral-7B-v1 & \ci{0.57}{0.47}{0.68} & \ci{0.36}{0.22}{0.49} & \ci{0.48}{0.39}{0.57} & \ci{0.57}{0.48}{0.65} & \ci{-0.30}{-0.37}{-0.23} & \ci{3.37}{3.30}{3.44} & \ci{0.94}{0.93}{0.96} & \ci{1.77}{1.66}{1.87} \\
SR-V2-Qwen3-0.6B & \ci{0.60}{0.54}{0.66} & \ci{0.55}{0.46}{0.62} & \ci{0.55}{0.49}{0.61} & \ci{0.57}{0.49}{0.64} & \ci{0.11}{0.02}{0.20} & \ci{2.77}{2.70}{2.84} & \ci{0.99}{0.98}{1.00} & \ci{1.75}{1.67}{1.83} \\
\bottomrule
\end{tabular}}
\caption{Table~\ref{tab:conditions} with 95\% intervals. The dagger marks the training-data overlap.}
\label{tab:conditions-full}
\end{table*}

\begin{table*}[htbp]
\centering
\footnotesize
\resizebox{\textwidth}{!}{
\begin{tabular}{lrrrrrrr}
\toprule
Model & Share & Generic prompt & Shown & Sibling & Sibling, generic prompt & Margin & Accuracy \\
\midrule
URM-LLaMa-3.1-8B & \ci{0.58}{0.51}{0.63} & 0.54 & \ci{0.48}{0.43}{0.52} & \ci{0.14}{0.08}{0.21} & \ci{0.07}{-0.02}{0.17} & \ci{2.63}{2.55}{2.71} & \ci{0.98}{0.98}{0.99} \\
Qwen3-8B & \ci{0.46}{0.39}{0.53} & 0.67 & \ci{0.43}{0.38}{0.49} & \ci{0.07}{0.01}{0.14} & \ci{-0.05}{-0.13}{0.02} & \ci{2.92}{2.83}{3.01} & \ci{0.94}{0.93}{0.96} \\
ArmoRM-Llama3-8B-v0.1 & \ci{0.49}{0.41}{0.56} & 0.79 & \ci{0.59}{0.53}{0.65} & \ci{0.11}{0.05}{0.18} & \ci{0.00}{-0.08}{0.08} & \ci{3.95}{3.87}{4.03} & \ci{0.99}{0.99}{0.99} \\
DPA-Mistral-7B-v1 & \ci{0.45}{0.37}{0.53} & 0.79 & \ci{0.59}{0.53}{0.65} & \ci{0.11}{0.05}{0.17} & \ci{-0.02}{-0.11}{0.06} & \ci{3.57}{3.50}{3.65} & \ci{0.99}{0.99}{1.00} \\
SR-Gemma-2-27B-v0.2 & \ci{0.64}{0.58}{0.69} & 0.50 & \ci{0.43}{0.39}{0.47} & \ci{0.13}{0.06}{0.19} & \ci{0.05}{-0.03}{0.13} & \ci{3.62}{3.52}{3.73} & \ci{0.98}{0.97}{0.99} \\
SR-Llama-3.1-8B-v0.2 & \ci{0.65}{0.59}{0.70} & 0.47 & \ci{0.41}{0.37}{0.46} & \ci{0.16}{0.09}{0.22} & \ci{0.08}{0.00}{0.16} & \ci{3.05}{2.96}{3.14} & \ci{0.98}{0.97}{0.99} \\
SR-V2-Llama-3.1-8B & \ci{0.55}{0.49}{0.60} &  & \ci{0.43}{0.38}{0.48} & \ci{0.09}{0.04}{0.15} &  &  &  \\
\bottomrule
\end{tabular}}
\caption{HelpSteer validation results for the eighteen released scorers: the between-prompt share under the original and under the generic prompt, the shown and sibling coefficients of the sibling-rating regression, and the mismatch penalty and match accuracy under four random wrong prompts. Blank cells are conditions that were not scored.}
\label{tab:hs1}
\end{table*}

\begin{table*}[htbp]
\centering
\small
\begin{tabular}{llrrr}
\toprule
Source & Model & share, raw & share, rank & rank minus raw \\
\midrule
human & human helpfulness & \ci{-0.01}{-0.11}{0.10} & \ci{-0.10}{-0.19}{-0.01} & -0.09 \\
model & URM-LLaMa-3.1-8B & \ci{0.34}{0.24}{0.43} & \ci{0.40}{0.31}{0.48} & 0.06 \\
model & Qwen3-8B & \ci{0.50}{0.41}{0.58} & \ci{0.58}{0.51}{0.64} & 0.08 \\
model & ArmoRM-Llama3-8B-v0.1 & \ci{0.51}{0.41}{0.60} & \ci{0.61}{0.55}{0.67} & 0.10 \\
model & DPA-Mistral-7B-v1 & \ci{0.57}{0.47}{0.68} & \ci{0.68}{0.61}{0.74} & 0.11 \\
model & SR-Gemma-2-27B-v0.2 & \ci{0.61}{0.53}{0.68} & \ci{0.63}{0.57}{0.70} & 0.02 \\
model & SR-Llama-3.1-8B-v0.2 & \ci{0.45}{0.36}{0.53} & \ci{0.50}{0.42}{0.57} & 0.05 \\
model & SR-V2-Llama-3.1-8B & \ci{0.44}{0.36}{0.52} & \ci{0.45}{0.36}{0.52} & 0.00 \\
model & SR-V2-Llama-3.2-1B & \ci{0.55}{0.48}{0.62} & \ci{0.55}{0.48}{0.62} & 0.00 \\
model & SR-V2-Llama-3.2-3B & \ci{0.49}{0.41}{0.57} & \ci{0.51}{0.43}{0.58} & 0.01 \\
model & SR-V2-Qwen3-0.6B & \ci{0.60}{0.54}{0.66} & \ci{0.59}{0.53}{0.65} & -0.01 \\
model & SR-V2-Qwen3-1.7B & \ci{0.55}{0.48}{0.62} & \ci{0.55}{0.48}{0.62} & 0.00 \\
model & SR-V2-Qwen3-4B & \ci{0.54}{0.46}{0.61} & \ci{0.54}{0.47}{0.61} & 0.00 \\
model & SR-V2-Qwen3-8B & \ci{0.53}{0.45}{0.60} & \ci{0.53}{0.46}{0.60} & 0.00 \\
model & Tulu-3-8B-RM & \ci{0.64}{0.57}{0.71} & \ci{0.64}{0.57}{0.70} & 0.00 \\
model & Llama-3.1-70B-Instruct & \ci{0.32}{0.22}{0.41} & \ci{0.46}{0.38}{0.53} & 0.14 \\
model & Llama-3.1-8B-Instruct & \ci{0.40}{0.30}{0.49} & \ci{0.52}{0.45}{0.59} & 0.12 \\
model & Llama-3.1-Nemotron-70B-Reward-HF & \ci{0.27}{0.17}{0.36} & \ci{0.30}{0.20}{0.38} & 0.03 \\
\bottomrule
\end{tabular}
\caption{Human and model between-prompt shares from raw scores and from pooled ranks on HelpSteer2 validation, with the rank-minus-raw difference, for seventeen released scorers.}
\label{tab:rank-hs2}
\end{table*}

\begin{table*}[htbp]
\centering
\small
\begin{tabular}{llrrr}
\toprule
Source & Model & share, raw & share, rank & rank minus raw \\
\midrule
human & human helpfulness & \ci{0.26}{0.20}{0.33} & \ci{0.26}{0.20}{0.33} & 0.00 \\
model & URM-LLaMa-3.1-8B & \ci{0.58}{0.51}{0.63} & \ci{0.58}{0.52}{0.63} & 0.00 \\
model & Qwen3-8B & \ci{0.46}{0.39}{0.53} & \ci{0.55}{0.49}{0.60} & 0.09 \\
model & ArmoRM-Llama3-8B-v0.1 & \ci{0.49}{0.41}{0.56} & \ci{0.52}{0.46}{0.58} & 0.03 \\
model & DPA-Mistral-7B-v1 & \ci{0.45}{0.37}{0.53} & \ci{0.54}{0.47}{0.59} & 0.08 \\
model & SR-Gemma-2-27B-v0.2 & \ci{0.64}{0.58}{0.69} & \ci{0.63}{0.58}{0.68} & -0.01 \\
model & SR-Llama-3.1-8B-v0.2 & \ci{0.65}{0.59}{0.70} & \ci{0.62}{0.57}{0.67} & -0.03 \\
model & SR-V2-Llama-3.1-8B & \ci{0.55}{0.49}{0.60} & \ci{0.56}{0.50}{0.61} & 0.01 \\
model & SR-V2-Llama-3.2-1B & \ci{0.56}{0.50}{0.61} & \ci{0.59}{0.54}{0.64} & 0.03 \\
model & SR-V2-Llama-3.2-3B & \ci{0.56}{0.50}{0.61} & \ci{0.57}{0.51}{0.62} & 0.01 \\
model & SR-V2-Qwen3-0.6B & \ci{0.55}{0.50}{0.60} & \ci{0.57}{0.52}{0.63} & 0.02 \\
model & SR-V2-Qwen3-1.7B & \ci{0.55}{0.49}{0.60} & \ci{0.56}{0.51}{0.62} & 0.02 \\
model & SR-V2-Qwen3-4B & \ci{0.54}{0.47}{0.59} & \ci{0.54}{0.49}{0.60} & 0.01 \\
model & SR-V2-Qwen3-8B & \ci{0.56}{0.50}{0.61} & \ci{0.57}{0.51}{0.62} & 0.01 \\
model & Tulu-3-8B-RM & \ci{0.61}{0.56}{0.66} & \ci{0.63}{0.58}{0.68} & 0.02 \\
model & Llama-3.1-70B-Instruct & \ci{0.54}{0.48}{0.60} & \ci{0.55}{0.49}{0.60} & 0.01 \\
model & Llama-3.1-8B-Instruct & \ci{0.59}{0.52}{0.66} & \ci{0.60}{0.54}{0.65} & 0.01 \\
model & Llama-3.1-Nemotron-70B-Reward-HF & \ci{0.57}{0.50}{0.62} & \ci{0.55}{0.49}{0.61} & -0.02 \\
\bottomrule
\end{tabular}
\caption{Human and model between-prompt shares from raw scores and from pooled ranks on HelpSteer validation, with the rank-minus-raw difference, for seventeen released scorers.}
\label{tab:rank-hs1}
\end{table*}

\begin{table*}[htbp]
\centering
\footnotesize
\resizebox{\textwidth}{!}{
\begin{tabular}{llrrr}
\toprule
Model & Set & calibration $R^2$ & calibrated, between-prompt variance & human, observed \\
\midrule
SR-V2-Qwen3-8B & HelpSteer2 & 0.20 & \ci{0.17}{0.14}{0.20} & \ci{-0.01}{-0.17}{0.16} \\
SR-Llama-3.1-8B-v0.2 & HelpSteer2 & 0.22 & \ci{0.16}{0.12}{0.20} & \ci{-0.01}{-0.17}{0.16} \\
Tulu-3-8B-RM & HelpSteer2 & 0.15 & \ci{0.16}{0.13}{0.18} & \ci{-0.01}{-0.17}{0.16} \\
ArmoRM-Llama3-8B-v0.1 & HelpSteer2 & 0.20 & \ci{0.16}{0.13}{0.20} & \ci{-0.01}{-0.17}{0.16} \\
URM-LLaMa-3.1-8B & HelpSteer2 & 0.36 & \ci{0.19}{0.13}{0.25} & \ci{-0.01}{-0.17}{0.16} \\
DPA-Mistral-7B-v1 & HelpSteer2 & 0.14 & \ci{0.13}{0.10}{0.16} & \ci{-0.01}{-0.17}{0.16} \\
\midrule
SR-V2-Qwen3-8B & HelpSteer & 0.22 & \ci{0.11}{0.09}{0.13} & \ci{0.22}{0.15}{0.28} \\
SR-Llama-3.1-8B-v0.2 & HelpSteer & 0.22 & \ci{0.12}{0.10}{0.14} & \ci{0.22}{0.15}{0.28} \\
Tulu-3-8B-RM & HelpSteer & 0.18 & \ci{0.09}{0.08}{0.11} & \ci{0.22}{0.15}{0.28} \\
ArmoRM-Llama3-8B-v0.1 & HelpSteer & 0.40 & \ci{0.16}{0.13}{0.21} & \ci{0.22}{0.15}{0.28} \\
URM-LLaMa-3.1-8B & HelpSteer & 0.27 & \ci{0.13}{0.11}{0.16} & \ci{0.22}{0.15}{0.28} \\
DPA-Mistral-7B-v1 & HelpSteer & 0.40 & \ci{0.15}{0.12}{0.19} & \ci{0.22}{0.15}{0.28} \\
\bottomrule
\end{tabular}}
\caption{The between-prompt variance of each model's score after a linear calibration to helpfulness cross-fitted by prompt, in squared rating units, with the calibration $R^2$ and the observed human between-prompt variance for comparison.}
\label{tab:calibrated}
\end{table*}

\begin{table*}[htbp]
\centering
\footnotesize
\resizebox{\textwidth}{!}{
\begin{tabular}{lllrrrr}
\toprule
Model & Set & Residualized on & between removed & within removed & share before & share after \\
\midrule
SR-V2-Qwen3-8B & HelpSteer2 & five ratings & \ci{0.12}{0.01}{0.22} & \ci{0.39}{0.30}{0.46} & 0.53 & 0.62 \\
SR-V2-Qwen3-8B & HelpSteer2 & log length & \ci{0.13}{0.03}{0.22} & \ci{0.10}{0.00}{0.19} & 0.53 & 0.52 \\
SR-V2-Qwen3-8B & HelpSteer2 & response probe & \ci{0.14}{0.05}{0.22} & \ci{0.05}{-0.06}{0.14} & 0.53 & 0.50 \\
SR-Llama-3.1-8B-v0.2 & HelpSteer2 & five ratings & \ci{0.06}{-0.07}{0.17} & \ci{0.38}{0.29}{0.45} & 0.45 & 0.55 \\
SR-Llama-3.1-8B-v0.2 & HelpSteer2 & log length & \ci{0.04}{-0.01}{0.10} & \ci{0.01}{-0.03}{0.04} & 0.45 & 0.44 \\
SR-Llama-3.1-8B-v0.2 & HelpSteer2 & response probe & \ci{0.07}{0.01}{0.13} & \ci{0.00}{-0.04}{0.05} & 0.45 & 0.43 \\
Tulu-3-8B-RM & HelpSteer2 & five ratings & \ci{0.20}{0.11}{0.28} & \ci{0.32}{0.19}{0.43} & 0.64 & 0.68 \\
\bottomrule
\end{tabular}}
\caption{The fraction of each core model's between-prompt and within-prompt variance removed by each adjustment, with the share before and after, on both validation sets. The response probe is HelpSteer2 only. A negative value means that the variance increased.}
\label{tab:components}
\end{table*}

\begin{table*}[htbp]
\centering
\footnotesize
\resizebox{\textwidth}{!}{
\begin{tabular}{llrrrrr}
\toprule
Model & Set & Concordance & Split halves & Difference variance & Share, original minus generic & Share, original minus length removed \\
\midrule
SR-V2-Qwen3-8B & HelpSteer2 & \ci{0.51}{0.44}{0.57} & \ci{0.27}{0.20}{0.33} & \ci{1.07}{0.95}{1.24} & \ci{0.01}{-0.09}{0.11} & \ci{0.01}{-0.03}{0.06} \\
SR-Llama-3.1-8B-v0.2 & HelpSteer2 & \ci{0.13}{0.04}{0.21} & \ci{0.02}{-0.06}{0.09} & \ci{1.80}{1.56}{2.16} & \ci{0.01}{-0.12}{0.13} & \ci{0.01}{-0.01}{0.04} \\
Tulu-3-8B-RM & HelpSteer2 & \ci{0.81}{0.77}{0.84} & \ci{0.48}{0.42}{0.55} & \ci{0.45}{0.39}{0.54} & \ci{0.09}{0.03}{0.15} & \ci{0.03}{-0.02}{0.09} \\
ArmoRM-Llama3-8B-v0.1 & HelpSteer2 & \ci{0.58}{0.52}{0.64} & \ci{0.34}{0.27}{0.41} & \ci{1.04}{0.87}{1.28} & \ci{-0.08}{-0.18}{0.03} & \ci{0.05}{-0.02}{0.12} \\
URM-LLaMa-3.1-8B & HelpSteer2 & \ci{0.14}{0.03}{0.23} & \ci{0.03}{-0.07}{0.12} & \ci{2.47}{1.95}{3.36} & \ci{-0.26}{-0.40}{-0.10} & \ci{0.02}{0.00}{0.06} \\
DPA-Mistral-7B-v1 & HelpSteer2 & \ci{0.73}{0.67}{0.78} & \ci{0.40}{0.32}{0.48} & \ci{0.69}{0.54}{0.89} & \ci{0.09}{-0.01}{0.19} & \ci{0.00}{-0.06}{0.07} \\
\midrule
SR-V2-Qwen3-8B & HelpSteer & \ci{0.11}{0.00}{0.21} & \ci{0.06}{-0.03}{0.15} & \ci{1.39}{1.25}{1.57} & \ci{-0.05}{-0.13}{0.02} & \ci{-0.02}{-0.03}{-0.01} \\
\bottomrule
\end{tabular}}
\caption{Changes in prompt means and between-prompt shares under the generic prompt and under length adjustment, per core model and rating set. Concordance is the Pearson correlation between each prompt's mean score under the original and under the generic prompt, and the split-half version uses disjoint responses for the two means. The difference variance is the variance of the per-prompt difference divided by the original between-prompt variance, so a ratio above one is possible. The share differences are original minus adjusted, with paired intervals.}
\label{tab:concordance}
\end{table*}

\begin{table*}[htbp]
\centering
\footnotesize
\resizebox{\textwidth}{!}{
\begin{tabular}{lrrrrrrr}
\toprule
Reward model & Pooled & Within prompt & Centered & Lower-rated kept, raw & Lower-rated kept, centered & Kept mean rating, raw & Kept mean rating, centered \\
\midrule
SR-V2-Qwen3-8B & \ci{0.40}{0.34}{0.45} & \ci{0.77}{0.72}{0.81} & \ci{0.58}{0.52}{0.64} & 0.21 & 0.11 & \ci{3.47}{3.40}{3.55} & \ci{3.44}{3.32}{3.54} \\
SR-Llama-3.1-8B-v0.2 & \ci{0.41}{0.36}{0.46} & \ci{0.75}{0.70}{0.79} & \ci{0.59}{0.52}{0.65} & 0.23 & 0.10 & \ci{3.40}{3.31}{3.49} & \ci{3.39}{3.27}{3.50} \\
Tulu-3-8B-RM & \ci{0.32}{0.27}{0.38} & \ci{0.73}{0.68}{0.77} & \ci{0.47}{0.39}{0.55} & 0.24 & 0.16 & \ci{3.42}{3.32}{3.51} & \ci{3.24}{3.12}{3.37} \\
ArmoRM-Llama3-8B-v0.1 & \ci{0.35}{0.29}{0.40} & \ci{0.73}{0.68}{0.77} & \ci{0.50}{0.42}{0.57} & 0.27 & 0.17 & \ci{3.38}{3.29}{3.47} & \ci{3.25}{3.13}{3.37} \\
URM-LLaMa-3.1-8B & \ci{0.50}{0.44}{0.54} & \ci{0.74}{0.70}{0.79} & \ci{0.59}{0.51}{0.65} & 0.21 & 0.14 & \ci{3.54}{3.46}{3.61} & \ci{3.30}{3.18}{3.42} \\
DPA-Mistral-7B-v1 & \ci{0.30}{0.24}{0.36} & \ci{0.70}{0.66}{0.75} & \ci{0.41}{0.32}{0.50} & 0.26 & 0.18 & \ci{3.35}{3.24}{3.45} & \ci{3.19}{3.04}{3.31} \\
SR-V2-Qwen3-0.6B & \ci{0.34}{0.28}{0.39} & \ci{0.73}{0.68}{0.77} & \ci{0.50}{0.43}{0.56} & 0.25 & 0.12 & \ci{3.35}{3.26}{3.44} & \ci{3.35}{3.23}{3.46} \\
\bottomrule
\end{tabular}}
\caption{HelpSteer2 validation agreement and top-30\% selection under raw and sibling-centered scores. Pooled is the Spearman correlation with the raw helpfulness rating, within prompt is the fraction of untied prompts on which the model orders the two responses as the raters did, centered is the pooled correlation after centering, and the selection columns give the fraction of kept responses rated below their sibling and the mean kept rating.}
\label{tab:cost}
\end{table*}

\begin{table*}[htbp]
\centering
\footnotesize
\setlength{\tabcolsep}{4pt}
\begin{tabular}{lrrrrrrrr}
\toprule
& \multicolumn{4}{c}{HelpSteer, independent ratings} & \multicolumn{4}{c}{HelpSteer2, sequential ratings} \\
\cmidrule(lr){2-5}\cmidrule(lr){6-9}
& \multicolumn{2}{c}{raw rating} & \multicolumn{2}{c}{centered rating} & \multicolumn{2}{c}{raw rating} & \multicolumn{2}{c}{centered rating} \\
\cmidrule(lr){2-3}\cmidrule(lr){4-5}\cmidrule(lr){6-7}\cmidrule(lr){8-9}
Reward model & raw score & centered & raw score & centered & raw score & centered & raw score & centered \\
\midrule
SR-V2-Qwen3-8B & 0.45 & 0.34 & 0.24 & 0.45 & 0.40 & 0.43 & 0.27 & 0.58 \\
SR-Llama-3.1-8B-v0.2 & 0.44 & 0.33 & 0.23 & 0.44 & 0.41 & 0.43 & 0.29 & 0.59 \\
Tulu-3-8B-RM & 0.39 & 0.31 & 0.22 & 0.42 & 0.32 & 0.35 & 0.19 & 0.47 \\
ArmoRM-Llama3-8B-v0.1 & 0.54 & 0.40 & 0.31 & 0.56 & 0.35 & 0.36 & 0.21 & 0.50 \\
URM-LLaMa-3.1-8B & 0.48 & 0.35 & 0.26 & 0.47 & 0.50 & 0.43 & 0.32 & 0.59 \\
DPA-Mistral-7B-v1 & 0.53 & 0.38 & 0.30 & 0.53 & 0.30 & 0.29 & 0.15 & 0.41 \\
SR-V2-Qwen3-0.6B & 0.41 & 0.33 & 0.25 & 0.46 & 0.34 & 0.36 & 0.22 & 0.50 \\
\bottomrule
\end{tabular}
\caption{The pooled Spearman correlations behind Figure~\ref{fig:centering}a and b, crossing the raw and the sibling-centered score with the raw and the sibling-centered helpfulness rating on each validation set. The centered entries are pooled Spearman correlations between sibling-centered scores and sibling-centered ratings.}
\label{tab:centering}
\end{table*}

\begin{table*}[htbp]
\centering
\footnotesize
\resizebox{\textwidth}{!}{
\begin{tabular}{lrrrrrr}
\toprule
Model & raw & leave-one-out & pool, 1 & pool, 12 & partial & shrinkage \\
\midrule
SR-V2-Qwen3-8B & 3.27 & \ci{-0.15}{-0.24}{-0.07} & \ci{-0.33}{-0.42}{-0.24} & \ci{-0.33}{-0.43}{-0.23} & \ci{0.00}{0.00}{0.00} & \ci{-0.31}{-0.41}{-0.22} \\
SR-Llama-3.1-8B-v0.2 & 3.24 & \ci{-0.14}{-0.24}{-0.06} & \ci{-0.32}{-0.41}{-0.23} & \ci{-0.26}{-0.38}{-0.17} & \ci{0.00}{0.00}{0.00} & \ci{-0.26}{-0.36}{-0.17} \\
Tulu-3-8B-RM & 3.20 & \ci{-0.15}{-0.23}{-0.07} & \ci{-0.20}{-0.28}{-0.12} & \ci{-0.15}{-0.25}{-0.08} & \ci{0.00}{-0.05}{0.04} & \ci{-0.15}{-0.24}{-0.07} \\
ArmoRM-Llama3-8B-v0.1 & 3.33 & \ci{-0.21}{-0.31}{-0.13} & \ci{-0.16}{-0.22}{-0.11} & \ci{-0.13}{-0.21}{-0.06} & \ci{0.00}{0.00}{0.00} & \ci{-0.13}{-0.20}{-0.06} \\
URM-LLaMa-3.1-8B & 3.28 & \ci{-0.19}{-0.27}{-0.09} & \ci{-0.19}{-0.26}{-0.12} & \ci{-0.14}{-0.22}{-0.07} & \ci{0.00}{-0.03}{0.02} & \ci{-0.14}{-0.21}{-0.07} \\
DPA-Mistral-7B-v1 & 3.30 & \ci{-0.21}{-0.30}{-0.12} & \ci{-0.08}{-0.14}{-0.04} & \ci{-0.05}{-0.13}{-0.01} & \ci{0.00}{0.00}{0.00} & \ci{-0.04}{-0.12}{0.00} \\
\bottomrule
\end{tabular}}
\caption{The mean helpfulness of the 30\% of HelpSteer responses kept by raw score, and the change in that mean under each centering method: leave-one-out sibling centering, an independent pool of one or twelve generated responses per prompt, partial centering with a weight chosen on held-out prompts, and the shrinkage baseline. Brackets are paired 95\% prompt-bootstrap intervals. This table is a separate run of the sweep with its own reference scoring and bootstrap. Its twelve-reference column differs from Table~\ref{tab:primary} by at most 0.016, and Section~\ref{sec:centering} reports the primary run's partial-centering choice.}
\label{tab:pool}
\end{table*}

\begin{table*}[htbp]
\centering
\footnotesize
\begin{tabular}{llrrrr}
\toprule
Model & Condition & Shown & Sibling & Between & Within \\
\midrule
SR-V2-Qwen3-8B & original & \ci{0.45}{0.39}{0.51} & \ci{0.05}{-0.01}{0.11} & \ci{0.50}{0.40}{0.61} & \ci{0.40}{0.35}{0.46} \\
SR-V2-Qwen3-8B & generic & \ci{0.30}{0.23}{0.36} & \ci{0.08}{0.02}{0.15} & \ci{0.38}{0.26}{0.50} & \ci{0.22}{0.15}{0.29} \\
SR-V2-Qwen3-8B & random wrong & \ci{0.16}{0.10}{0.22} & \ci{0.03}{-0.03}{0.10} & \ci{0.19}{0.10}{0.29} & \ci{0.13}{0.04}{0.21} \\
SR-V2-Qwen3-8B & nearest wrong & \ci{0.13}{0.07}{0.19} & \ci{0.03}{-0.03}{0.09} & \ci{0.16}{0.05}{0.27} & \ci{0.10}{0.04}{0.16} \\
\midrule
SR-Llama-3.1-8B-v0.2 & original & \ci{0.47}{0.42}{0.53} & \ci{0.01}{-0.04}{0.07} & \ci{0.49}{0.40}{0.59} & \ci{0.46}{0.40}{0.52} \\
SR-Llama-3.1-8B-v0.2 & generic & \ci{0.08}{0.02}{0.15} & \ci{-0.02}{-0.08}{0.05} & \ci{0.07}{-0.04}{0.17} & \ci{0.10}{0.02}{0.18} \\
SR-Llama-3.1-8B-v0.2 & random wrong & \ci{0.02}{-0.05}{0.08} & \ci{-0.04}{-0.11}{0.02} & \ci{-0.03}{-0.14}{0.08} & \ci{0.06}{-0.02}{0.14} \\
\bottomrule
\end{tabular}
\caption{The sibling-rating regression on helpfulness, HelpSteer2 validation, under the original prompt, the generic prompt, four random wrong prompts and the four nearest wrong prompts by TF-IDF: the shown and sibling coefficients and the between-prompt and within-prompt slopes, in reward standard deviations per rating standard deviation.}
\label{tab:swaphs2}
\end{table*}

\begin{table*}[htbp]
\centering
\footnotesize
\begin{tabular}{llrrrr}
\toprule
Model & Condition & Shown & Sibling & Between & Within \\
\midrule
SR-V2-Qwen3-8B & original & \ci{0.44}{0.39}{0.49} & \ci{0.11}{0.05}{0.17} & \ci{0.59}{0.50}{0.69} & \ci{0.39}{0.34}{0.44} \\
SR-V2-Qwen3-8B & generic & \ci{-0.01}{-0.07}{0.05} & \ci{0.03}{-0.04}{0.11} & \ci{0.04}{-0.10}{0.18} & \ci{-0.02}{-0.08}{0.03} \\
SR-V2-Qwen3-8B & random wrong & \ci{-0.14}{-0.20}{-0.07} & \ci{0.02}{-0.07}{0.11} & \ci{-0.11}{-0.26}{0.05} & \ci{-0.15}{-0.22}{-0.08} \\
SR-V2-Qwen3-8B & nearest wrong & \ci{-0.07}{-0.12}{-0.01} & \ci{0.09}{0.00}{0.18} & \ci{0.06}{-0.09}{0.21} & \ci{-0.11}{-0.16}{-0.05} \\
\midrule
SR-Llama-3.1-8B-v0.2 & original & \ci{0.41}{0.37}{0.46} & \ci{0.16}{0.09}{0.23} & \ci{0.63}{0.52}{0.75} & \ci{0.34}{0.29}{0.39} \\
SR-Llama-3.1-8B-v0.2 & generic & \ci{-0.10}{-0.16}{-0.04} & \ci{0.08}{0.00}{0.16} & \ci{0.01}{-0.13}{0.15} & \ci{-0.13}{-0.20}{-0.07} \\
SR-Llama-3.1-8B-v0.2 & random wrong & \ci{-0.20}{-0.26}{-0.14} & \ci{0.04}{-0.05}{0.13} & \ci{-0.15}{-0.31}{0.02} & \ci{-0.22}{-0.27}{-0.16} \\
\bottomrule
\end{tabular}
\caption{The sibling-rating regression on helpfulness, HelpSteer validation, with the conditions and units of Table~\ref{tab:swaphs2}.}
\label{tab:swaphs1}
\end{table*}

\begin{table*}[htbp]
\centering
\footnotesize
\begin{tabular}{llrrr}
\toprule
Model & Prompt variant & Random donors, margin & Margin & Accuracy \\
\midrule
SR-V2-Qwen3-8B & sub-question & \ci{3.39}{3.28}{3.49} & \ci{1.15}{1.03}{1.27} & \ci{0.77}{0.72}{0.82} \\
SR-V2-Qwen3-8B & superset question & \ci{3.42}{3.32}{3.51} & \ci{1.00}{0.92}{1.09} & \ci{0.95}{0.92}{0.97} \\
SR-V2-Qwen3-8B & different domain & \ci{3.41}{3.30}{3.52} & \ci{2.67}{2.52}{2.81} & \ci{0.99}{0.98}{1.00} \\
SR-Llama-3.1-8B-v0.2 & sub-question & \ci{3.83}{3.71}{3.96} & \ci{1.26}{1.13}{1.40} & \ci{0.72}{0.66}{0.77} \\
SR-Llama-3.1-8B-v0.2 & superset question & \ci{3.86}{3.73}{3.98} & \ci{1.19}{1.09}{1.30} & \ci{0.95}{0.91}{0.97} \\
SR-Llama-3.1-8B-v0.2 & different domain & \ci{3.89}{3.77}{4.02} & \ci{3.44}{3.27}{3.62} & \ci{0.98}{0.96}{1.00} \\
Tulu-3-8B-RM & sub-question & \ci{1.30}{1.24}{1.36} & \ci{0.70}{0.63}{0.76} & \ci{0.77}{0.71}{0.81} \\
\bottomrule
\end{tabular}
\caption{Mismatch penalties and match accuracies for three prompt variants, each on 170 to 199 single-turn HelpSteer2 validation prompts, with the random-donor penalty on the same prompts. Penalties are in original-condition standard deviations and accuracies are proportions.}
\label{tab:graded}
\end{table*}

\begin{table*}[htbp]
\centering
\footnotesize
\resizebox{\textwidth}{!}{
\begin{tabular}{lcccccccccccc}
\toprule
& \multicolumn{6}{c}{HelpSteer2} & & & & & & \\
\cmidrule(lr){2-7}
Model & real & generic & random donors & nearest donors & related tasks & paraphrases & HelpSteer & Reddit & history & leaderboard & template variants & reference pool \\
\midrule
URM-LLaMa-3.1-8B & yes & yes & yes & yes & yes & yes & yes & yes & yes & yes & yes & yes \\
Qwen3-8B & yes & yes & yes & yes & -- & -- & yes & -- & yes & -- & -- & -- \\
ArmoRM-Llama3-8B-v0.1 & yes & yes & yes & yes & yes & yes & yes & yes & yes & yes & yes & yes \\
DPA-Mistral-7B-v1 & yes & yes & yes & yes & yes & yes & yes & yes & yes & yes & yes & yes \\
SR-Gemma-2-27B-v0.2 & yes & yes & yes & yes & -- & -- & yes & -- & yes & -- & -- & -- \\
SR-Llama-3.1-8B-v0.2 & yes & yes & yes & yes & yes & yes & yes & yes & yes & yes & yes & yes \\
\bottomrule
\end{tabular}}
\caption{Scoring coverage by model, rating set and condition. A dash marks a condition that was not scored.}
\label{tab:coverage}
\end{table*}

\begin{table*}[htbp]
\centering
\footnotesize
\resizebox{\textwidth}{!}{
\begin{tabular}{lrrrr}
\toprule
Model & $\tau$, raw & $\tau$, centered & difference & assignments differing from full coverage, of 20 \\
\midrule
URM-LLaMa-3.1-8B & \ci{0.53}{0.44}{0.67} & \ci{0.55}{0.45}{0.71} & \ci{0.02}{-0.07}{0.13} & 11 \\
ArmoRM-Llama3-8B-v0.1 & \ci{0.40}{0.31}{0.55} & \ci{0.43}{0.33}{0.58} & \ci{0.03}{-0.06}{0.12} & 8 \\
DPA-Mistral-7B-v1 & \ci{0.50}{0.35}{0.69} & \ci{0.60}{0.44}{0.79} & \ci{0.10}{-0.05}{0.28} & 12 \\
SR-Llama-3.1-8B-v0.2 & \ci{0.80}{0.80}{1.00} & \ci{0.80}{0.80}{1.00} & \ci{0.00}{-0.03}{0.05} & 0 \\
SR-V2-Qwen3-8B & \ci{0.87}{0.80}{0.95} & \ci{0.88}{0.80}{0.97} & \ci{0.01}{-0.05}{0.06} & 7 \\
Tulu-3-8B-RM & \ci{0.92}{0.84}{1.00} & \ci{0.98}{0.86}{1.00} & \ci{0.06}{-0.04}{0.11} & 8 \\
\midrule
SR-V2-Qwen3-0.6B & \ci{0.99}{0.80}{1.00} & \ci{0.99}{0.80}{1.00} & \ci{0.00}{-0.05}{0.08} & 1 \\
SR-V2-Llama-3.1-8B & \ci{0.80}{0.71}{0.94} & \ci{0.80}{0.79}{1.00} & \ci{0.00}{0.00}{0.08} & 0 \\
SR-V2-Llama-3.2-1B & \ci{0.70}{0.57}{0.87} & \ci{0.75}{0.65}{0.91} & \ci{0.05}{-0.05}{0.17} & 10 \\
SR-V2-Llama-3.2-3B & \ci{0.80}{0.69}{0.92} & \ci{0.80}{0.75}{0.97} & \ci{0.00}{-0.02}{0.10} & 0 \\
SR-V2-Qwen3-1.7B & \ci{1.00}{0.80}{1.00} & \ci{1.00}{0.80}{1.00} & \ci{0.00}{0.00}{0.02} & 0 \\
SR-V2-Qwen3-4B & \ci{0.89}{0.82}{0.98} & \ci{0.92}{0.82}{1.00} & \ci{0.03}{-0.04}{0.08} & 11 \\
\bottomrule
\end{tabular}}
\caption{Kendall's tau between each reward model's ranking of the five policies under unequal prompt coverage and the 70B judge's full-coverage ranking, over twenty prompt-subset assignments, for raw and for centered mean rewards. The last column counts the assignments in which the raw ranking differs from the same reward model's full-coverage ranking.}
\label{tab:leaderboard}
\end{table*}

\begin{table*}[htbp]
\centering
\footnotesize
\resizebox{\textwidth}{!}{
\begin{tabular}{lrr}
\toprule
Model & Mean absolute shift, last user turn only & Mean absolute shift, assistant turns omitted \\
\midrule
URM-LLaMa-3.1-8B & \ci{1.08}{0.95}{1.21} & \ci{0.42}{0.36}{0.47} \\
Qwen3-8B & \ci{0.63}{0.53}{0.73} & \ci{0.43}{0.36}{0.50} \\
ArmoRM-Llama3-8B-v0.1 & \ci{0.82}{0.74}{0.92} & \ci{0.58}{0.52}{0.64} \\
DPA-Mistral-7B-v1 & \ci{0.70}{0.62}{0.81} & \ci{1.40}{1.26}{1.55} \\
SR-Gemma-2-27B-v0.2 & \ci{0.68}{0.62}{0.75} & \ci{0.39}{0.35}{0.42} \\
SR-Llama-3.1-8B-v0.2 & \ci{0.78}{0.69}{0.87} & \ci{0.46}{0.41}{0.51} \\
SR-V2-Llama-3.1-8B & \ci{0.69}{0.61}{0.77} & \ci{0.44}{0.39}{0.49} \\
SR-V2-Llama-3.2-1B & \ci{0.73}{0.65}{0.81} & \ci{0.57}{0.51}{0.63} \\
SR-V2-Llama-3.2-3B & \ci{0.62}{0.55}{0.70} & \ci{0.48}{0.43}{0.53} \\
SR-V2-Qwen3-0.6B & \ci{0.61}{0.54}{0.68} & \ci{0.35}{0.31}{0.39} \\
SR-V2-Qwen3-1.7B & \ci{0.63}{0.56}{0.69} & \ci{0.38}{0.34}{0.43} \\
SR-V2-Qwen3-4B & \ci{0.64}{0.56}{0.72} & \ci{0.32}{0.28}{0.35} \\
SR-V2-Qwen3-8B & \ci{0.64}{0.57}{0.72} & \ci{0.32}{0.28}{0.36} \\
Tulu-3-8B-RM & \ci{0.57}{0.51}{0.62} & \ci{0.60}{0.54}{0.66} \\
Llama-3.1-70B-Instruct & \ci{0.69}{0.57}{0.83} & \ci{0.28}{0.23}{0.34} \\
Llama-3.1-8B-Instruct & \ci{0.70}{0.59}{0.82} & \ci{0.38}{0.32}{0.44} \\
Llama-3.1-Nemotron-70B-Reward-HF & \ci{1.07}{0.95}{1.19} & \ci{0.44}{0.39}{0.50} \\
\bottomrule
\end{tabular}}
\caption{Two history ablations on the 163 multi-turn HelpSteer2 validation prompts, for seventeen released scorers: the final response is scored with only the last user turn, or with the earlier assistant turns omitted. Entries are the mean absolute shift of the response's score in original-condition standard deviations.}
\label{tab:history}
\end{table*}

\begin{table*}[htbp]
\centering
\footnotesize
\resizebox{\textwidth}{!}{
\begin{tabular}{lrrrrr}
\toprule
Model & Full history & Last user turn only & Assistant turns omitted & Single-turn prompts & Last user turn minus single-turn \\
\midrule
URM-LLaMa-3.1-8B & \ci{0.45}{0.27}{0.60} & \ci{0.71}{0.61}{0.79} & \ci{0.53}{0.37}{0.67} & \ci{0.28}{0.18}{0.39} & \ci{0.43}{0.29}{0.57} \\
Qwen3-8B & \ci{0.59}{0.46}{0.70} & \ci{0.46}{0.30}{0.60} & \ci{0.58}{0.46}{0.69} & \ci{0.40}{0.28}{0.51} & \ci{0.06}{-0.13}{0.24} \\
ArmoRM-Llama3-8B-v0.1 & \ci{0.67}{0.53}{0.78} & \ci{0.69}{0.59}{0.77} & \ci{0.68}{0.55}{0.79} & \ci{0.43}{0.31}{0.54} & \ci{0.26}{0.11}{0.40} \\
DPA-Mistral-7B-v1 & \ci{0.82}{0.73}{0.89} & \ci{0.60}{0.44}{0.72} & \ci{0.65}{0.51}{0.76} & \ci{0.36}{0.22}{0.50} & \ci{0.24}{0.04}{0.43} \\
SR-Gemma-2-27B-v0.2 & \ci{0.65}{0.55}{0.75} & \ci{0.67}{0.56}{0.77} & \ci{0.64}{0.52}{0.75} & \ci{0.38}{0.26}{0.51} & \ci{0.29}{0.12}{0.45} \\
SR-Llama-3.1-8B-v0.2 & \ci{0.57}{0.42}{0.70} & \ci{0.67}{0.55}{0.76} & \ci{0.62}{0.47}{0.75} & \ci{0.36}{0.25}{0.46} & \ci{0.30}{0.15}{0.45} \\
SR-V2-Llama-3.1-8B & \ci{0.50}{0.35}{0.62} & \ci{0.62}{0.50}{0.72} & \ci{0.60}{0.46}{0.71} & \ci{0.32}{0.19}{0.42} & \ci{0.30}{0.14}{0.46} \\
\bottomrule
\end{tabular}}
\caption{The between-prompt share under the full history, under the two history ablations of Table~\ref{tab:history}, and on the single-turn prompts, with the difference between the last-user-turn share and the single-turn share.}
\label{tab:history-share}
\end{table*}

\begin{table*}[htbp]
\centering
\footnotesize
\resizebox{\textwidth}{!}{
\begin{tabular}{llrrrrrrr}
\toprule
Arm & Seed & Share & Generic prompt & Margin & Accuracy & AUC & HelpSteer share & HelpSteer swap \\
\midrule
0.6B regression & 0 & 0.33 & 0.50 & 2.60 & 0.96 & 0.982 & 0.47 & 0.01 \\
0.6B regression & 1 & 0.29 & 0.44 & 2.68 & 0.95 & 0.981 & 0.46 & 0.00 \\
0.6B regression & 2 & 0.32 & 0.49 & 2.48 & 0.95 & 0.980 & 0.42 & -0.02 \\
0.6B regression & 3 & 0.34 & 0.46 & 2.50 & 0.97 & 0.985 & 0.48 & 0.04 \\
0.6B regression & 4 & 0.46 & 0.59 & 2.27 & 0.95 & 0.979 & 0.45 & 0.02 \\
0.6B Bradley--Terry & 0 & 0.35 & 0.42 & 2.02 & 0.87 & 0.951 & 0.67 & -0.06 \\
0.6B Bradley--Terry & 1 & 0.34 & 0.40 & 1.96 & 0.92 & 0.969 & 0.68 & -0.07 \\
0.6B Bradley--Terry & 2 & 0.54 & 0.43 & 1.66 & 0.90 & 0.955 & 0.64 & -0.03 \\
0.6B Bradley--Terry & 3 & 0.37 & 0.39 & 2.10 & 0.92 & 0.969 & 0.76 & -0.07 \\
0.6B Bradley--Terry & 4 & 0.56 & 0.53 & 1.65 & 0.91 & 0.962 & 0.60 & -0.08 \\
\midrule
1.7B regression & 0 & 0.33 & 0.41 & 2.79 & 0.98 & 0.993 & 0.43 & 0.07 \\
1.7B regression & 1 & 0.34 & 0.43 & 2.49 & 0.97 & 0.986 & 0.45 & 0.09 \\
1.7B regression & 2 & 0.29 & 0.45 & 2.84 & 0.99 & 0.994 & 0.45 & 0.07 \\
1.7B regression & 3 & 0.35 & 0.37 & 2.72 & 0.98 & 0.994 & 0.44 & 0.07 \\
1.7B regression & 4 & 0.34 & 0.44 & 2.91 & 0.98 & 0.994 & 0.48 & 0.09 \\
1.7B Bradley--Terry & 0 & 0.54 & 0.48 & 1.86 & 0.94 & 0.979 & 0.81 & -0.01 \\
1.7B Bradley--Terry & 1 & 0.64 & 0.62 & 2.03 & 0.98 & 0.992 & 0.69 & 0.04 \\
1.7B Bradley--Terry & 2 & 0.48 & 0.53 & 2.20 & 0.95 & 0.982 & 0.70 & -0.04 \\
1.7B Bradley--Terry & 3 & 0.57 & 0.56 & 2.25 & 0.97 & 0.988 & 0.56 & -0.06 \\
1.7B Bradley--Terry & 4 & 0.46 & 0.46 & 1.67 & 0.92 & 0.969 & 0.77 & 0.01 \\
\bottomrule
\end{tabular}}
\caption{Per-seed results for the two main matched objectives at 0.6B and 1.7B. The HelpSteer2 columns give the share under the original and the generic prompt and the random-prompt mismatch margin and accuracy, and the last columns the HelpSteer share and sibling coefficient.}
\label{tab:matched}
\end{table*}

\begin{table*}[htbp]
\centering
\footnotesize
\resizebox{\textwidth}{!}{
\begin{tabular}{llrrrrrr}
\toprule
Size & Arm & Margin & Share & Generic prompt & Match accuracy & Pair accuracy & HelpSteer share \\
\midrule
0.6B & regression & \ci{2.51}{2.27}{2.68} & \ci{0.35}{0.29}{0.46} & 0.49 & 0.96 & 0.69 & 0.46 \\
0.6B & Bradley--Terry & \ci{1.88}{1.65}{2.10} & \ci{0.43}{0.34}{0.56} & 0.44 & 0.91 & 0.73 & 0.67 \\
0.6B & regression, Bradley--Terry configuration & \ci{2.29}{2.04}{2.52} & \ci{0.40}{0.36}{0.43} & 0.49 & 0.95 & 0.69 & 0.46 \\
0.6B & Bradley--Terry, regression configuration & \ci{2.01}{1.69}{2.19} & \ci{0.43}{0.36}{0.45} & 0.40 & 0.89 & 0.71 & 0.67 \\
0.6B & rating-margin Bradley--Terry & \ci{2.34}{2.12}{2.89} & \ci{0.34}{0.29}{0.39} & 0.44 & 0.96 & 0.73 & 0.65 \\
0.6B & binarized regression & \ci{0.71}{0.49}{0.97} & \ci{0.34}{0.25}{0.49} & 0.30 & 0.60 & 0.65 & 0.57 \\
0.6B & gap-only regression & \ci{2.30}{2.06}{2.64} & \ci{0.47}{0.35}{0.54} & 0.48 & 0.94 & 0.72 & 0.61 \\
0.6B & shuffled-mean regression & \ci{2.20}{2.09}{2.41} & \ci{0.21}{0.15}{0.25} & 0.35 & 0.90 & 0.70 & 0.53 \\
\midrule
1.7B & regression & \ci{2.75}{2.49}{2.91} & \ci{0.33}{0.29}{0.35} & 0.42 & 0.98 & 0.74 & 0.45 \\
1.7B & Bradley--Terry & \ci{2.00}{1.67}{2.25} & \ci{0.54}{0.46}{0.64} & 0.53 & 0.95 & 0.75 & 0.71 \\
1.7B & regression, Bradley--Terry configuration & \ci{2.40}{1.77}{2.78} & \ci{0.37}{0.34}{0.45} & 0.50 & 0.95 & 0.72 & 0.53 \\
1.7B & Bradley--Terry, regression configuration & \ci{2.06}{1.88}{2.17} & \ci{0.41}{0.37}{0.45} & 0.44 & 0.96 & 0.77 & 0.60 \\
1.7B & rating-margin Bradley--Terry & \ci{2.60}{2.24}{3.00} & \ci{0.49}{0.43}{0.59} & 0.51 & 0.97 & 0.75 & 0.69 \\
1.7B & binarized regression & \ci{1.95}{1.79}{2.03} & \ci{0.13}{0.11}{0.16} & 0.31 & 0.95 & 0.75 & 0.65 \\
1.7B & gap-only regression & \ci{2.38}{2.27}{2.44} & \ci{0.46}{0.44}{0.48} & 0.45 & 0.97 & 0.77 & 0.60 \\
1.7B & shuffled-mean regression & \ci{2.91}{2.77}{3.04} & \ci{0.20}{0.16}{0.22} & 0.35 & 0.98 & 0.76 & 0.55 \\
\bottomrule
\end{tabular}}
\caption{Every matched training condition and the two cross-configuration controls at both sizes. Entries are means over five seeds, and the brackets after the margin and the share are the minimum and maximum over seeds. The other columns give the generic-prompt share, the match accuracy, the held-out pair accuracy and the HelpSteer share.}
\label{tab:arms}
\end{table*}

\begin{table*}[htbp]
\centering
\footnotesize
\resizebox{\textwidth}{!}{
\begin{tabular}{lrrrr}
\toprule
Model & Length, all prompts & Length, single-turn & Length, single-turn, generic prompt & Topic, balanced \\
\midrule
URM-LLaMa-3.1-8B & \ci{-0.29}{-0.43}{-0.15} & \ci{-0.39}{-0.56}{-0.23} & \ci{-0.11}{-0.30}{0.08} & \ci{0.02}{-0.12}{0.17} \\
Qwen3-8B & \ci{-0.41}{-0.55}{-0.26} & \ci{-0.28}{-0.43}{-0.12} & \ci{-0.32}{-0.48}{-0.16} & \ci{0.12}{-0.03}{0.26} \\
ArmoRM-Llama3-8B-v0.1 & \ci{-0.27}{-0.42}{-0.12} & \ci{-0.33}{-0.50}{-0.17} & \ci{-0.53}{-0.70}{-0.35} & \ci{0.04}{-0.11}{0.19} \\
DPA-Mistral-7B-v1 & \ci{-0.62}{-0.76}{-0.47} & \ci{-0.31}{-0.47}{-0.16} & \ci{-0.39}{-0.55}{-0.23} & \ci{0.06}{-0.08}{0.21} \\
SR-Gemma-2-27B-v0.2 & \ci{-0.73}{-0.88}{-0.58} & \ci{-0.02}{-0.18}{0.13} & \ci{-0.16}{-0.35}{0.01} & \ci{0.09}{-0.06}{0.25} \\
SR-Llama-3.1-8B-v0.2 & \ci{-0.40}{-0.54}{-0.26} & \ci{-0.17}{-0.34}{0.01} & \ci{-0.03}{-0.22}{0.15} & \ci{0.10}{-0.05}{0.24} \\
SR-V2-Llama-3.1-8B & \ci{-0.46}{-0.61}{-0.32} & \ci{-0.19}{-0.35}{-0.04} & \ci{-0.40}{-0.56}{-0.24} & \ci{0.06}{-0.08}{0.21} \\
\bottomrule
\end{tabular}}
\caption{Standardized mean reward differences between prompt subsets on HelpSteer2 validation for seventeen released scorers. The length contrasts are the longer half minus the shorter half of the prompts, on all prompts, on single-turn prompts, and on single-turn prompts under the generic prompt. The topic contrast uses two balanced splits. The last row gives the human rating difference.}
\label{tab:shift}
\end{table*}

\begin{table*}[htbp]
\centering
\footnotesize
\resizebox{\textwidth}{!}{
\begin{tabular}{lrrrrrr}
\toprule
Reward model & random & TF-IDF & embedding & length-matched & related task & Paraphrase shift, mean absolute \\
\midrule
SR-V2-Qwen3-8B & 3.16 / 0.99 & 3.09 / 0.97 & 3.04 / 0.97 & 3.14 / 0.99 & 2.16 / 0.97 & 0.18 [0.16, 0.20] \\
SR-Llama-3.1-8B-v0.2 & 3.60 / 0.98 & 3.38 / 0.97 & 3.31 / 0.96 & 3.62 / 0.99 & 2.55 / 0.97 & 0.23 [0.21, 0.26] \\
Tulu-3-8B-RM & 1.09 / 0.88 & 1.19 / 0.90 & 1.16 / 0.88 & 1.06 / 0.88 & 1.08 / 0.96 & 0.14 [0.12, 0.15] \\
ArmoRM-Llama3-8B-v0.1 & 3.65 / 0.99 & 3.24 / 0.97 & 3.16 / 0.96 & 3.65 / 0.99 & 1.90 / 0.94 & 0.17 [0.15, 0.19] \\
URM-LLaMa-3.1-8B & 3.25 / 0.99 & 3.11 / 0.98 & 3.12 / 0.98 & 3.25 / 0.98 & 2.16 / 0.96 & 0.20 [0.17, 0.22] \\
DPA-Mistral-7B-v1 & 3.37 / 0.94 & 2.94 / 0.91 & 2.93 / 0.93 & 3.35 / 0.95 & 1.77 / 0.96 & 0.12 [0.10, 0.13] \\
SR-V2-Qwen3-0.6B & 2.77 / 0.99 & 2.71 / 0.97 & 2.66 / 0.97 & 2.77 / 0.99 & 1.75 / 0.94 & 0.19 [0.17, 0.21] \\
\bottomrule
\end{tabular}}
\caption{Mismatch margins and match accuracies under random, TF-IDF-nearest, embedding-nearest, length-matched and related-task transplants, and the mean absolute reward shift under three paraphrases. Margins and shifts are in original-condition standard deviations and accuracies are proportions.}
\label{tab:transplants}
\end{table*}

\begin{table*}[htbp]
\centering
\footnotesize
\resizebox{\textwidth}{!}{\input{tables/appendix_template_variants}}
\caption{The random-prompt mismatch margin and the between-prompt share of each core model on HelpSteer2 validation with the chat template's start marker, end marker or both removed. The difference columns are variant minus default.}
\label{tab:templates}
\end{table*}

\begin{table*}[htbp]
\centering
\footnotesize
\resizebox{\textwidth}{!}{
\begin{tabular}{lrrrr}
\toprule
Reward model & Shown & Sibling & Sibling, adjusted & Between-prompt share of the reward \\
\midrule
Skywork-Reward-V2-Qwen3-8B & 0.45 [0.39, 0.51] & 0.05 [-0.01, 0.11] & 0.03 [-0.02, 0.09] & 0.53 [0.45, 0.60] \\
Skywork-Reward-Llama-3.1-8B-v0.2 & 0.47 [0.41, 0.53] & 0.01 [-0.04, 0.07] & 0.01 [-0.05, 0.07] & 0.45 [0.36, 0.54] \\
Llama-3.1-Tulu-3-8B-RM & 0.40 [0.33, 0.46] & 0.10 [0.03, 0.16] & 0.08 [0.02, 0.15] & 0.64 [0.57, 0.71] \\
ArmoRM-Llama3-8B-v0.1 & 0.45 [0.38, 0.52] & 0.09 [0.02, 0.15] & 0.06 [0.00, 0.13] & 0.51 [0.42, 0.60] \\
URM-LLaMa-3.1-8B & 0.60 [0.54, 0.66] & 0.07 [0.02, 0.13] & 0.06 [0.00, 0.11] & 0.34 [0.24, 0.43] \\
RewardModel-Mistral-7B-for-DPA-v1 & 0.38 [0.30, 0.45] & 0.11 [0.04, 0.17] & 0.09 [0.02, 0.16] & 0.57 [0.47, 0.68] \\
Skywork-Reward-V2-Qwen3-0.6B & 0.39 [0.33, 0.44] & 0.06 [0.00, 0.11] & 0.04 [-0.02, 0.09] & 0.60 [0.54, 0.66] \\
\bottomrule
\end{tabular}}
\caption{The sibling-rating regression on helpfulness on HelpSteer2 validation (1,038 responses): the shown and sibling coefficients, the sibling coefficient after adjustment for the shown response's complexity and verbosity ratings, and the between-prompt share.}
\label{tab:swap}
\end{table*}

\begin{table*}[htbp]
\centering
\footnotesize
\resizebox{\textwidth}{!}{
\begin{tabular}{lrrrrr}
\toprule
Reward model & Help. & Corr. & Coh. & Compl. & Verb. \\
\midrule
SR-V2-Qwen3-8B & 0.05 [-0.01, 0.11] (0.151) & 0.05 [0.00, 0.11] (0.152) & -0.03 [-0.09, 0.02] (1.000) & -0.03 [-0.10, 0.05] (1.000) & 0.06 [-0.01, 0.12] (0.071) \\
SR-Llama-3.1-8B-v0.2 & 0.01 [-0.04, 0.07] (0.614) & 0.02 [-0.04, 0.07] (0.546) & -0.08 [-0.14, -0.02] (0.019) & -0.04 [-0.12, 0.04] (1.000) & 0.09 [0.03, 0.16] (0.018) \\
Tulu-3-8B-RM & 0.10 [0.03, 0.16] (0.012) & 0.10 [0.04, 0.17] (0.002) & 0.00 [-0.06, 0.05] (1.000) & -0.03 [-0.11, 0.04] (1.000) & 0.11 [0.05, 0.17] ($<$0.001) \\
ArmoRM-Llama3-8B-v0.1 & 0.09 [0.02, 0.15] (0.029) & 0.11 [0.04, 0.17] (0.006) & 0.01 [-0.04, 0.07] (1.000) & -0.01 [-0.10, 0.08] (1.000) & 0.10 [0.03, 0.16] (0.011) \\
URM-LLaMa-3.1-8B & 0.07 [0.02, 0.13] (0.027) & 0.08 [0.02, 0.13] (0.020) & -0.01 [-0.06, 0.04] (1.000) & 0.03 [-0.05, 0.12] (1.000) & 0.11 [0.04, 0.18] (0.011) \\
DPA-Mistral-7B-v1 & 0.11 [0.04, 0.17] (0.012) & 0.13 [0.06, 0.20] (0.002) & 0.03 [-0.03, 0.09] (1.000) & -0.03 [-0.11, 0.05] (1.000) & 0.10 [0.04, 0.16] (0.008) \\
\bottomrule
\end{tabular}}
\caption{The sibling coefficient per rating attribute on HelpSteer2 validation, with 95\% intervals in brackets and, in parentheses, the $p$-value Holm-adjusted over the six core models within each attribute.}
\label{tab:attr}
\end{table*}

\begin{table*}[htbp]
\centering
\footnotesize
\resizebox{\textwidth}{!}{\input{tables/appendix_strata}}
\caption{The between-prompt share of each core model on all HelpSteer2 validation prompts, on single-turn prompts, on the 398 clean prompts whose first turn appears in neither the Skywork preference sets nor HelpSteer, and on the training split, with the generic-prompt share on single-turn prompts and the correlation across prompts between an empty reply's score and the original prompt mean.}
\label{tab:strata}
\end{table*}

\begin{table*}[htbp]
\centering
\footnotesize
\resizebox{\textwidth}{!}{\input{tables/appendix_rb_subsets}}
\caption{Pairwise accuracy of the six core reward models on the four RewardBench sections and the RewardBench 2 subsets.}
\label{tab:rbsub}
\end{table*}

\end{document}